\documentclass[journal]{IEEEtran}
\usepackage{cite}
\usepackage{graphicx}
\usepackage{amsmath}
\usepackage{amssymb}
\usepackage[pagebackref,breaklinks,colorlinks]{hyperref}
\usepackage[capitalize]{cleveref}
\usepackage{multirow}
\usepackage{booktabs}

\ifCLASSINFOpdf
\else
\fi
\begin{document}
\title{A bioinspired three-stage model for camouflaged object detection}

\author{Tianyou Chen, Jin Xiao, Xiaoguang Hu, Guofeng Zhang, Shaojie Wang}


\markboth{Journal of \LaTeX\ Class Files,~Vol.~14, No.~8, August~2015}%
{Shell \MakeLowercase{\textit{et al.}}: Bare Demo of IEEEtran.cls for IEEE Journals}



\maketitle

\begin{abstract}
Camouflaged objects are typically assimilated into their backgrounds and exhibit fuzzy boundaries. The complex environmental conditions and the high intrinsic similarity between camouflaged targets and their surroundings pose significant challenges in accurately locating and segmenting these objects in their entirety. While existing methods have demonstrated remarkable performance in various real-world scenarios, they still face limitations when confronted with difficult cases, such as small targets, thin structures, and indistinct boundaries.
Drawing inspiration from human visual perception when observing images containing camouflaged objects, we propose a three-stage model that enables coarse-to-fine segmentation in a single iteration. Specifically, our model employs three decoders to sequentially process subsampled features, cropped features, and high-resolution original features. This proposed approach not only reduces computational overhead but also mitigates interference caused by background noise.
Furthermore, considering the significance of multi-scale information, we have designed a multi-scale feature enhancement module that enlarges the receptive field while preserving detailed structural cues. Additionally, a boundary enhancement module has been developed to enhance performance by leveraging boundary information. Subsequently, a mask-guided fusion module is proposed to generate fine-grained results by integrating coarse prediction maps with high-resolution feature maps.
Our network surpasses state-of-the-art CNN-based counterparts without unnecessary complexities. Upon acceptance of the paper, the source code will be made publicly available at https://github.com/clelouch/BTSNet.
\end{abstract}


% \fi
\IEEEpeerreviewmaketitle

\section{Introduction}

\IEEEPARstart{C}{amouflage} is widespread in nature, as wildlife animals often adapt their patterns and colors to blend into their surroundings or mimic other objects for concealment purposes \cite{BM-COD}. Through the use of camouflage, prey animals can effectively hinder the detection or recognition by their predators, thereby enhancing their protection\cite{CODhistory}. Furthermore, the concept of camouflaged objects extends beyond wildlife, encompassing entities that closely resemble the background or are heavily obscured, such as polyps or soldiers utilizing camouflage techniques. Within this context, the field of camouflaged object detection (COD) has garnered significant research attention, enabling a range of crucial applications including lung infection segmentation\cite{INFNet}, polyp segmentation\cite{PraNet}, species identification\cite{CODSpecies}, and military camouflage pattern design\cite{militaryCOD}.

\begin{figure}[htbp]
    \centering
    \includegraphics[width=1\linewidth]{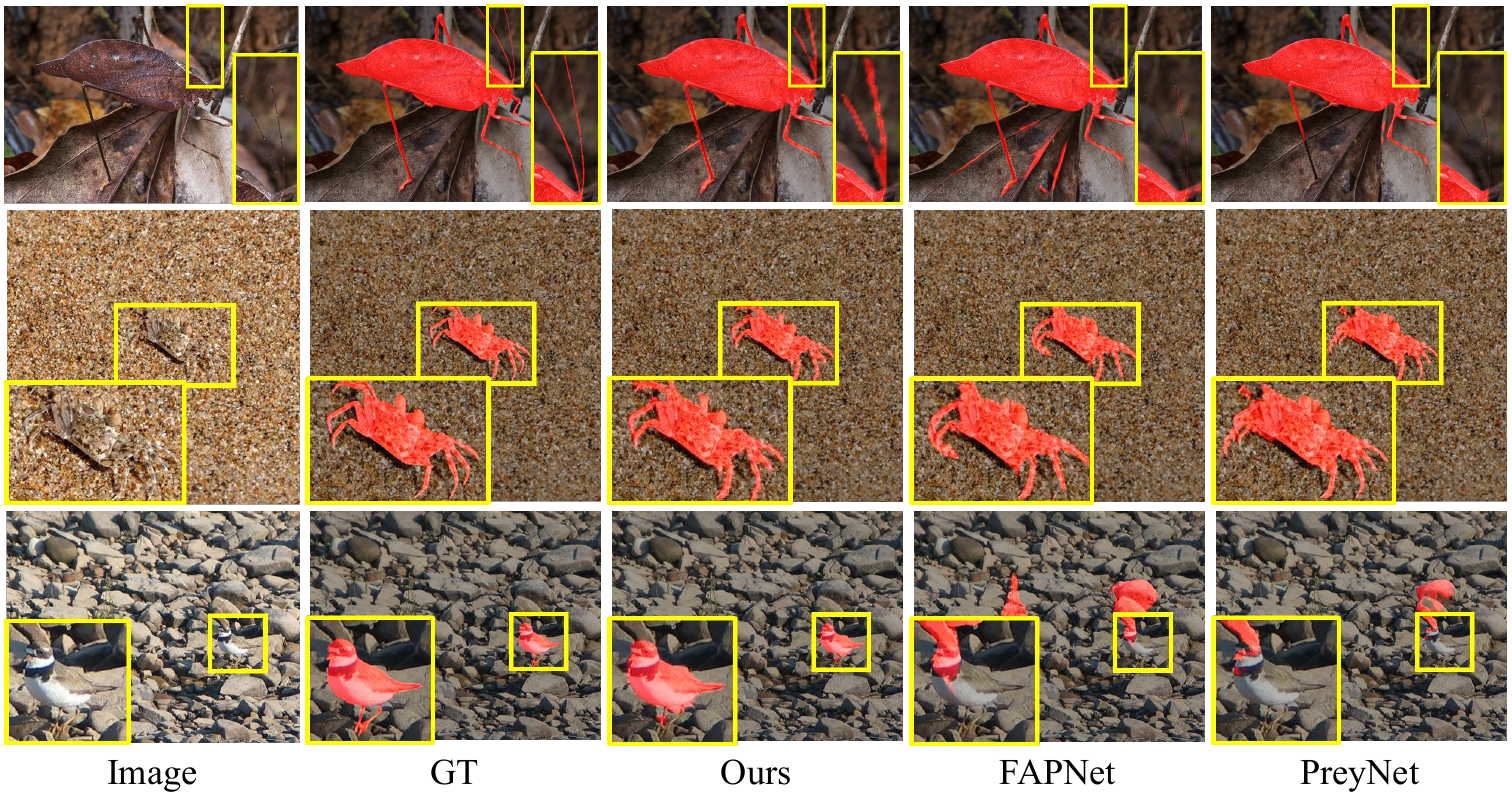}
    \caption{Several visual examples of camouflaged object detection in several highly challenging cases (e.g., long and thin structures, low contrast, small targets). The left (1st row) or right (2rd and 3th) yellow box areas are the enlarged regions cropped from the original image. Compared with the state-of-the-art CNN-based models (i.e., FAPNet \cite{FAPNet} and PreyNet \cite{PreyNet}, our method preserves better details and is competent to distinguish the entire target from the background.}
    \label{fig:intro}
\end{figure}

Recently, with the advancement of large-scale benchmark datasets\cite{MirrorNet,SINet,LSR}, a plethora of COD \cite{SINet,ERRNet,CubeNet,BGNet-IJCAI,PreyNet} methods have been proposed and have exhibited exceptional performance in diverse and intricate scenarios. In general, the detection process of these models can be summarized into three steps: 1) resizing the input image to a reduced resolution, 2) leveraging a pretrained backbone to extract features at multiple levels, and 3) utilizing a decoder to generate the final result. Despite the effectiveness of this widely adopted framework in addressing various challenging scenarios, there exist certain overlooked flaws that significantly hinder its overall performance.

Concretely, in the first step, the majority of COD models employ low-resolution images (e.g., $352 \times 352$ \cite{PFNet, SINetV2,FAPNet}, $416 \times 416$ \cite{BGNet-IJCAI}) as their input. However, camouflaged objects can be small in size and exhibit fuzzy boundaries. Although the subsampling process reduces training time and computational and memory overhead, it comes at the cost of losing substantial structural information, leading to a decline in performance. A possible remedy for this issue is the utilization of high-resolution images. Nevertheless, as highlighted in \cite{Segtran,TransTrans}, CNN-based methods have an effective receptive field size much smaller than the theoretical value. Consequently, solely using larger samples would make it challenging for the models to capture large targets accurately. Furthermore, in the last two steps, existing models rely on multiple convolution blocks to process the entire image sequentially. However, a significant portion of the image consists of redundant areas for COD. These extensive background regions not only squander computational resources but also introduce significant noise. As a result, the models may erroneously identify some background objects as camouflaged targets.

In this paper, we propose the Bioinspired Three-Stage Network (BTSNet) as a solution to address these issues. BTSNet draws inspiration from the human perceptual behavior when examining images containing camouflaged targets. Given an indistinct image, the observer's search process can be broadly divided into three sequential steps: 1) scanning the image to identify suspicious regions, 2) zooming in on each identified region to verify the presence of camouflaged objects, and 3) zooming out to determine the precise locations of all targets within the input image. We emulate this process and devise a novel framework that facilitates the generation of detailed outcomes through a coarse-to-fine approach.

Specifically, we propose a bifurcated backbone network that incorporates the initial two stages of the original ResNet-50 \cite{ResNet} model as the stem branch. Additionally, we employ the remaining three stages to construct two leaf branches with identical structures. In contrast to \cite{CPD}, we introduce a $2\times 2$ maxpooling layers to downsample the output of the stem branch. Subsequently, this downsampled output is fed into the first leaf branch. Features extracted from the first leaf branch are then aggregated by the first decoder to generate a coarse prediction map. Importantly, the primary focus of the first decoder is to precisely locate the target. Hence, it becomes imperative to capture multi-scale information to effectively characterize camouflaged objects of varying sizes. To address this issue, we propose the Multi-scale Feature Enhancement Module (MFEM). MFEM comprises multiple branches, each performing convolution operations within an individual scale space. Furthermore, the output of the preceding branch is combined with the input using element-wise addition before being passed to the subsequent branch to repeat this process. By systematically reducing the kernel size of the pooling layers, we can successfully extract multi-scale cues and significantly expand the receptive field size while retaining important structural information.

After obtaining the initial output of the first decoder, the corresponding foreground features are extracted from the stem branch through cropping and resized to a predetermined resolution. Subsequently, the second leaf branch utilizes these cropped features as input to generate more refined results, while being unaffected by background regions. To improve the overall performance, we introduce the Boundary Enhancement Module (BEM) in this process. The BEM integrates boundary information with the input feature through channel-wise concatenation. Furthermore, attention mechanisms are employed to enhance the representation capability. Through the incorporation of boundary information, the BEM effectively mitigates the problem of blurred contours.

The cropping and resizing operations conducted prior to the application of the second decoder have the potential to negatively impact the final performance. Therefore, in the third decoder, we incorporate the stem branch feature and the output of the second decoder as inputs in order to recover the omitted details. As emphasized in \cite{DSS,BINet,PoolNet}, low-level features (specifically, stem branch features) preserve ample details but are prone to background noises. Consequently, a simple integration of features with the generated mask may result in performance degradation. To address this issue, we propose the Mask-Guided Fusion Module (MGFM). The MGFM initially employs the Split-Fusion Module (SFM) for integration. The SFM divides the input feature into multiple groups, each of which is combined with the mask through channel-wise concatenation. By sequentially processing the multiple groups of features, we are able to effectively suppress background noises. Subsequently, the MGFM introduces boundary information to further enhance the performance. Consequently, we can generate fine-grained results with well-defined contours.

To validate the superiority of our proposed schema and the effectiveness of its key components, we conducted extensive experiments on three benchmark datasets. The experimental results demonstrate that the proposed schema not only generates high-quality results but also significantly reduces computational and memory overhead. Specifically, BTSNet surpasses state-of-the-art CNN-based counterparts by a substantial margin across six universally accepted evaluation metrics. Additionally, BTSNet achieves a processing speed of 50 frames per second (FPS) for $704 \times 704$ inputs with a batch size of 60, utilizing a single NVIDIA Titan XP GPU.

In summary, the main contributions of this paper are five-fold:

\begin{itemize}
    \item Inspired by human behavior when searching for camouflaged objects in vague images, a novel schema for COD is proposed. The proposed schema demonstrates the ability to leverage detailed structural information effectively, resulting in superb results while minimizing computational and memory overhead.
    \item The Multi-scale Feature Enhancement Module is designed to enhance the representation ability of the input feature. This module not only extracts multi-scale information but also preserves finer detailed cues, thereby improving the overall performance of the model.
    \item To incorporate boundary information effectively and propagate useful knowledge to shallower decoder stages, a Boundary Enhancement Module is proposed. This module facilitates the preservation of contour cues, leading to a significant boost in COD performance.
    \item A Mask-Guided Fusion Module is developed to generate fine-grained prediction maps by integrating semantic information from existing estimated target regions and complementary details from low-level side-output features. By exploiting feature complementarity fully, this module accurately locates targets and substantially eliminates ambiguous regions.
    \item Extensive experiments are conducted on three benchmark datasets. The experimental results validate the superiority of the proposed novel schema and the effectiveness of the key modules.
\end{itemize}

\section{related works}
\subsection{Camouflaged object detection}
Similar to many other computer vision tasks (e.g., salient object detection\cite{BASNet,BINet}, object detection\cite{FasterRCNN}), early traditional COD methods mainly rely on hand-crafted features to spot the camouflaged targets from their high-similarity surroundings\cite{CODTrad1,CODtrad2,CODtrad3,CODtrad4}. However, as pointed out in many previous researches \cite{SINetV2,BgNet-KBS}, these approaches are limited to only using low-level features to locate the targets, which makes these methods unable to distinguish camouflaged objects in complex scenarios, e.g., low contrast, occlusions, and indefinable boundaries.

Similar to many other computer vision tasks (e.g., salient object detection\cite{BASNet,BINet}, object detection\cite{FasterRCNN}), early traditional COD methods primarily rely on hand-crafted features to identify camouflaged targets within their highly similar surroundings. However, as highlighted in numerous prior research studies\cite{SINetV2,BgNet-KBS}, these approaches are constrained by their utilization of only low-level features for target localization. Consequently, these methods face limitations in effectively discerning camouflaged objects within complex scenarios characterized by factors such as low contrast, occlusions, and ambiguous boundaries.

Recently, with the development of finely annotated large-scale datasets\cite{SINet,MirrorNet,LSR}, numerous deep-learning-based COD methods have been proposed, surpassing their traditional counterparts by a significant margin. Fan et al. (SINet) \cite{SINet} propose a straightforward yet effective framework comprising two main modules for searching and detecting camouflaged targets. Met et al. \cite{PFNet} (PFNet) employ a positioning module to locate potential targets from a global perspective and design a focus module to iteratively refine the coarse prediction map. Sun et al.(C$^2$FNet) \cite{C2FNet} develop an attention-induced cross-level fusion module to aggregate features from multiple levels, followed by the utilization of a dual-branch global context module to exploit global contextual cues. Ren et al. (TANet) \cite{TANet} devise multiple texture-aware refinement modules to generate texture-aware features, thereby enhancing the segmentation process by amplifying the subtle texture differences between the camouflaged target and its surroundings. Lv et al. (LSR) \cite{LSR} incorporate ranking information and propose a ranking-based model for COD. Zhuge et al. (CubeNet) \cite{CubeNet} introduce the concept of $\chi$ connection and include a sub edge decoder between two square fusion decoders to effectively model the weak contours of the objects. Zhai et al. (DTCNet) \cite{DTCNet} develop a deep model that leverages the spatial organization of textons in both background and foreground areas as informative cues for accurate COD. Chou et al. (PINet)  \cite{PINet} devise a cascaded decamouflage module for target detection and obtaining complete segmentation results.

In addition to the aforementioned methods, supplementary information has recently been incorporated to enhance performance. Li et al. (FindNet) \cite{FindNet} introduce global contour and local patterns as crucial cues to facilitate the detection process. Chen et al. (BgNet) \cite{BgNet-KBS} explicitly exploit the complementary information between camouflaged regions and their corresponding boundaries, enabling the generation of prediction maps with finer details. He et al. (ELDNet) \cite{ELDNet} have designed a novel framework for progressively refining boundary likelihood maps, which are then utilized to guide the feature fusion of concealed targets. Sun et al. (BGNet) \cite{BGNet-IJCAI} utilize informative object-related edge semantics to emphasize object structure and improve the performance of COD. 

Despite the exceptional performance achieved by these methods, it is worth noting that most existing COD models are trained and tested on low-resolution (352$\times$352\cite{ERRNet,TINet,PraNet,SINetV2},416$\times$ 416\cite{BGNet-IJCAI}) images. The subsampling process prior to model input leads to the loss of structural information. Consequently, the subtle differences between the camouflaged targets and their surroundings become more ambiguous, significantly degrading performance. In contrast, we propose a novel COD schema that fully exploits contextual information and local details in high-resolution input images, with minimal additional computational and memory overhead.

\subsection{Multi-scale feature extraction}
It has been noted in numerous previous studies \cite{MINet,PoolNet,u2net,BASNet} that capturing objects of varying sizes requires the incorporation of multi-scale information. To address this issue, several modules have been proposed to enhance the representation ability and improve performance by extracting multi-scale information.

Zhang et al. \cite{BMPM} introduce the multi-scale context-aware feature extraction module, which employs four 3$\times$3 dilated convolutional layers with different dilation rates to effectively capture multi-scale context information. Qin et al. (U$^2$Net) \cite{u2net} adopt a U-shaped architecture and propose a novel residual U-block to extract intra-stage multi-scale features. Wu et al. (CPD) \cite{CPD} make modifications to the receptive field block\cite{RFB}. In the n-th branch of the modified module, a (2n-1)$\times$(2n-1) convolutional layer followed by a 3$\times$3 dilated convolutional layer with a dilation rate of (2n-1) is utilized to expand the receptive field size and exploit multi-scale context information. The efficacy of the aforementioned modules has been demonstrated in various approaches related to COD \cite{SINetV2,C2FNet,BgNet-KBS}.

Compared to the convolution operation, the pooling operation has fewer parameters and higher efficiency. Recently, several pooling-based modules have been proposed to extract multi-scale context cues. Ji et al. (FSNet) \cite{FSNet} introduce the Pyramid Pooling Module (PPM) \cite{PPM} to capture targets of various sizes. PPM consists of multiple branches, each containing an adaptive average pooling layer followed by a 1$\times$1 convolutional layer. However, the intermediate feature maps produced by the adaptive average pooling layers lack detailed information, making PPM more suitable for processing low-resolution features.

In contrast, Liu et al. \cite{PoolNet} propose the Feature Aggregation Module (FAM). Similar to PPM, FAM also comprises multiple branches. However, FAM uses an average pooling layer in each branch to better preserve the structural information. Building upon the concept of FAM, we propose a pooling-based module named Multi-scale Feature Enhancement Module (MFEM) to characterize multi-scale information. In contrast to FAM, which calculates intermediate features in parallel using multiple branches, the branches in MFEM are utilized in series. This sequential utilization allows the output of each preceding branch to propagate to the next branch, enabling further enhancement. By progressively reducing the size of the pooling layers, we can significantly expand the receptive field size while preserving structural information.

\begin{figure*}[htbp]
    \centering
    \includegraphics[width=1\linewidth]{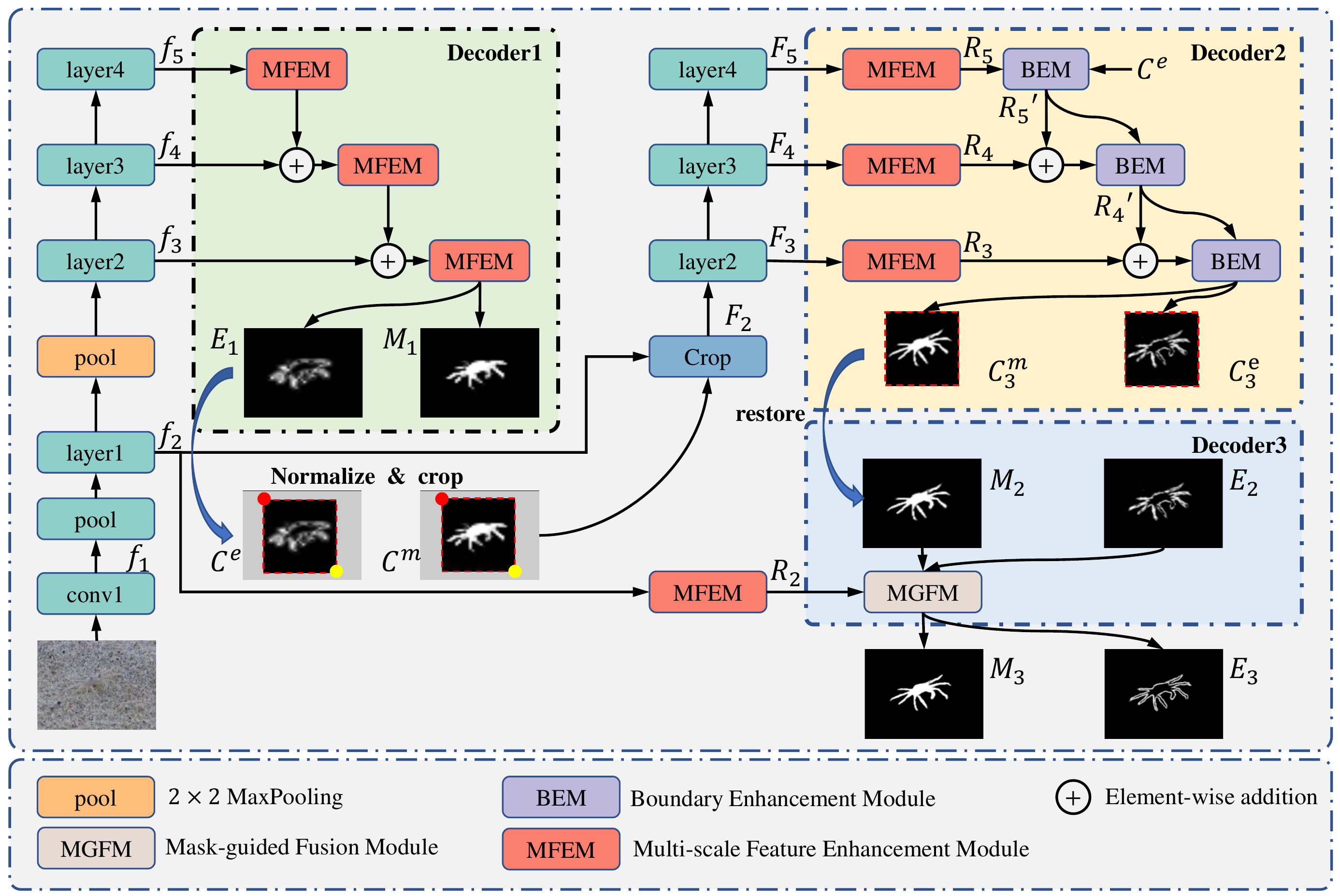}
    \caption{The overall pipeline of the BTSNet (Best viewed in color).}
    \label{fig:btsnet}
\end{figure*}

\subsection{Multi-stage detection}
Employing multiple decoders to perform coarse-to-fine detection is a common practice in the field. In the early CNN-based methods, several decoders with identical architecture are primarily utilized to enhance the segmentation results. For instance, Wei et al. (F$^3$Net) \cite{F3Net} introduce the Cascaded Feedback Decoder (CFD), which comprises multiple sub-decoders with independent parameters but sharing the same architecture. This design allows for iterative refinement of multi-level features, leading to improved performance. Similarly, Chen et al. (AFNet) \cite{AFNet-RGBD} propose the cascaded feature interweaved decoder to exploit the complementary information across multi-level features and iteratively refine them for producing more precise segmentation maps.

Afterwards, certain methodologies have indicated that employing decoders with distinct architectures at different stages can potentially yield improved performance. In their work, Fan et al. (SINetV2) \cite{SINetV2} adopt a neighbor connection decoder to integrate features for generating a coarse result. Subsequently, they employ multiple group-reversal attention modules to generate the refined prediction map. In a similar vein, Chen et al. (BgNet) \cite{BgNet-KBS} employ a simplified decoder for target localization. Subsequently, the coarse result, along with the boundary prediction map, is fed into the second decoder to generate a segmentation result with enhanced boundary delineation.

It is worth noting that the aforementioned methods extract hierarchical features from the entire image, which results in computational overhead due to the inclusion of background regions and negatively impacts the final performance. Therefore, a potential solution could be to assign less importance to the background regions and instead focus on the foreground areas. Xu et al. (PA-KRN) \cite{PAKRN} propose a coarse locating module to approximate the regions containing the target. Based on this coarse result, an attention-based sampler \cite{AttSampler} is employed to emphasize informative regions in the input image. Subsequently, the magnified image is fed into the fine segmenting module to generate the final results. Similarly, Jia et al. (SegMaR) \cite{SegMaR} also utilize the sampler \cite{AttSampler} to integrate segmentation, magnification, and iterative processes in a multi-stage fashion.

Benefiting from the sampler, these methods exhibit improved refinement and enrichment of details, particularly for small camouflaged targets. However, it is important to note that these methods require repeated extraction of encoder features and generation of full-resolution prediction maps, leading to decreased running speed and limited practical applications. In contrast, our multi-stage detection framework does not rely on the non-parametric sampler, enabling an end-to-end model. Consequently, our model demonstrates higher efficiency, and the optimization of the multi-stage training process becomes easier.
 
\section{Methods}\label{sec:methods}

\subsection{Overview}
The overall framework of BTSNet is depicted in \cref{fig:btsnet}. BTSNet consists of a bifurcated backbone network and three decoders. Given an input image with a spatial resolution of $H \times W$, the first branch of the backbone extracts five hierarchical features, denoted as $\{f_i, i=1,2,3,4,5\}$. It is worth noting that we add a $2\times 2$ maxpooling layer after obtaining $f_2$. As emphasized in \cite{Segtran,TransTrans}, the effective receptive field size of a CNN is smaller than its theoretical value. The inclusion of a pooling layer facilitates the model in fully leveraging global context information, without introducing additional parameters.

We discard $f_1$ since low-level features largely increase the inference time while contributing less to the final performance \cite{CPD}. Thus, the spatial resolution of the extracted features can be calculated as follows:
\begin{equation}
    \begin{cases}
        $$\frac{H}{2^{i}} \times \frac{W}{2^{i}} &i\leq2 \\
        \frac{H}{2^{i+1}} \times \frac{W}{2^{i+1}} &i>2$$
    \end{cases}
\end{equation}

\begin{figure*}[htbp]
    \centering
    \includegraphics[width=1\linewidth]{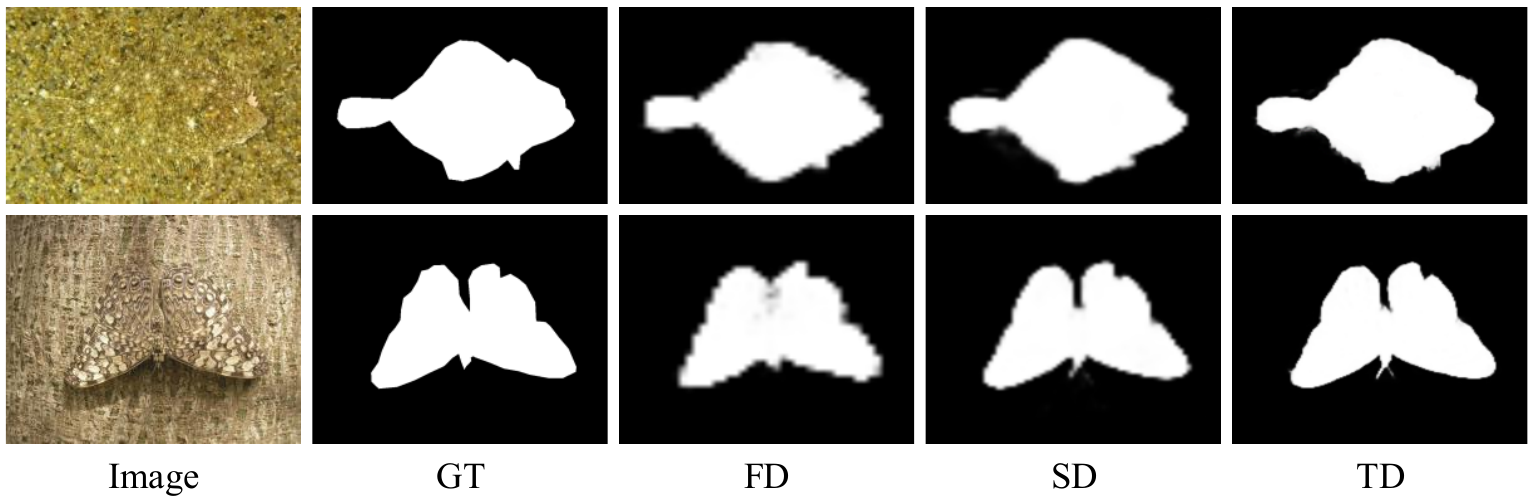}
    \caption{Results of different decoders (Best viewed in zoomed-in). FD: first decoder. SD: second decoder. TD: third decoder.}
    \label{fig:result-decoders}
\end{figure*}

Then, the three high-level features (i.e., $f_3$, $f_4$, $f_5$) are utilized as inputs to the first decoder in order to generate a preliminary map. The primary objective of the initial decoder is to extract multi-scale information to identify all potential targets. In order to simplify the process, we employ a U-shaped architecture. This procedure can be described as follows:

\begin{equation}
    r_5 = MFEM(f_5),
\end{equation}
\begin{equation}
    r_4 = MFEM(f_4 + r_5),
\end{equation}
\begin{equation}
    r_3 = MFEM(f_3 + r_4),
\end{equation}
where $MFEM$ indicates the MFEM module. After obtaining the refined feature $r_3$, we use a $3\times 3$ convolutional layer to produce a coarse result $M_1$. Following \cite{BgNet-KBS}, the corresponding boundary prediction map $E_1$ can be calculated as follows:
\begin{equation}
    E_1 = {\rm abs}(P^a_3(M_1) - M_1)
\end{equation}
where abs denotes the absolute value function, $P^a_3$ represents a $3\times 3$ average pooling layer with 1 stride. 

After obtaining $M_1$ and $E_1$, we use a normalization function to ensure that the two coarse prediction maps are within the range of [0, 1]. We note that even background regions contain non-zero elements. To reduce the negative impacts caused by background noise, a binarization process is performed by applying a threshold value of 0.5. The formulation for this process is as follows:
\begin{equation}
    M_1^n = f_{minmax}(M_1),E_1^n = f_{minmax}(E_1),
\end{equation}
\begin{equation}
    M_1^b(i,j)=\begin{cases}
        1,&M_1^n(i,j)>0.5 \\
        0,&M_1^n(i,j)\le 0.5
    \end{cases}
\end{equation}
\begin{equation}
    E_1^b(i,j)=\begin{cases}
        1,&E_1^n(i,j)>0.5 \\
        0,&E_1^n(i,j)\le 0.5
    \end{cases}
\end{equation}
where $f_{minmax}$ represents the normalization function, $M_1^n$ and $E_1^n$ denote the normalized results, $M_1^b$ and $E_1^b$ are binary masks, $M_1^b(i,j)$ and $E_1^b(i,j)$ are the pixel values of $M_1^b$ and $E_1^b$.

Inspired by Faster RCNN \cite{FasterRCNN} and Mask RCNN\cite{MaskRCNN}, we generate the bounding box of the located target and crop the corresponding features from $f_2$ to exclude background regions. Specifically, we first upsample $M_1^b$ to match the size of $f_2$ and calculate the sum of the upsampled result along the height/width dimension. Notably, non-zero elements can be identified at both ends of the resulting one-dimensional vector. Thus, we can get the initial bounding box ($x_{min}$,$y_{min}$,$x_{max}$,$y_{max}$). The center coordinates of the box are computed as ($x_c$=($x_{min}$+$x_{max}$)/2, $y_c$=($y_{min}$+$y_{max}$)/2). Since $M_1^b$ may contain incomplete parts, we expand the initial bounding box to encompass a larger area. Therefore, the final bounding box can be represented as ($x_c-r\times l/2$, $y_c-r\times l/2$, $x_c+r \times l/2$, $y_c+r \times l/2$), where $l=max(y_{max}-y_{min}, x_{max}-x_{min})$ and $r$ denotes the expansion ratio. Based on the final bounding box, we crop a feature map from $f_2$. The cropped feature is denoted as $F_2$ and is empirically resized to $120 \times 120$ before being inputted into the second branch of the backbone to generate semantic-enhanced features $F_3, F_4, F_5$. Unlike $f_i (i=3,4,5)$, $F_i (i=3,4,5$) is specifically designed to emphasize foreground regions. It is noteworthy that previous methods employ $352 \times 352$ images as inputs. The resolutions of the three deepest features are $44\times 44$, $22 \times 22$, and $11 \times 11$. In contrast, the second branch of our backbone extracts foreground features with larger sizes (i.e., $60 \times 60$, $30 \times 30$, and $15 \times 15$). Thus, we can mitigate the negative impacts caused by background regions and preserve finer foreground details.

The second decoder takes the cropped boundary prediction map $C^e$ and the foreground features $F_3, F_4, F_5$ as inputs. It is important to note that $C^e$ is obtained by cropping from $E_1^n$ based on the bounding box. In this stage, we initially utilize MFEMs to expand the receptive fields and excavate multi-scale contextual information, thereby enhancing the performance. Subsequently, the refined features, along with the boundary prediction map, are inputted to the BEM to generate precise outcomes. This process can be formulated as follows:
\begin{equation}
    R_i = MFEM(F_i),i=3,4,5,
\end{equation}
\begin{equation}
    C^m_i, C^e_i, R_i' = \begin{cases}
        BEM(R_i, R_{i+1}', C^e_{i+1}),&i=3,4,\\
        BEM(R_5, C^e),&i=5,\\
    \end{cases}
\end{equation}
where $C^m_i$ denotes the prediction map, $C^e_i$ represents the boundary prediction map, $R_i'$ is a enhanced feature. $C^m_3$ and $C^e_3$ are then resized to the same size as the bounding box. We create an image with the same size as $f_2$ and assign zero value to all pixels in the image. Afterward, we can obtain $M_2$ and $E_2$ by mapping $C^m_3$ and $C^e_3$ into the bounding box region of the created image. 

As illustrated in \cref{fig:result-decoders}, targets in $M_2$ often exhibit jagged boundaries, which can be attributed to low-resolution features (i.e., $F_3$) and the application of multiple resizing operations. To achieve fine-grained segmentation results with smoother boundaries, the third decoder takes $f_2$, $M_2$, and $E_2$ as inputs to generate the final results. The process can be described as:
\begin{equation}
    M_3, E_3 = MGFM(f_2, M_2, E_2),
\end{equation}
where $MGFM$ is the Mask-Guide Fusion Module, $M_3$ and $E_3$ are the prediction results.

As done in many previous methods\cite{BBSNet,DSS}, we adopt deep supervision strategy to train the whole model in an end-to-end manner. Without using any post-processing technique (e.g., CRF \cite{CRF}), our proposed BTSNet generates fine-grained results with a real-time inference speed of 50 FPS on a single NVIDIA Titan XP GPU.

\begin{figure}[htbp]
    \centering
    \includegraphics[width=1\linewidth]{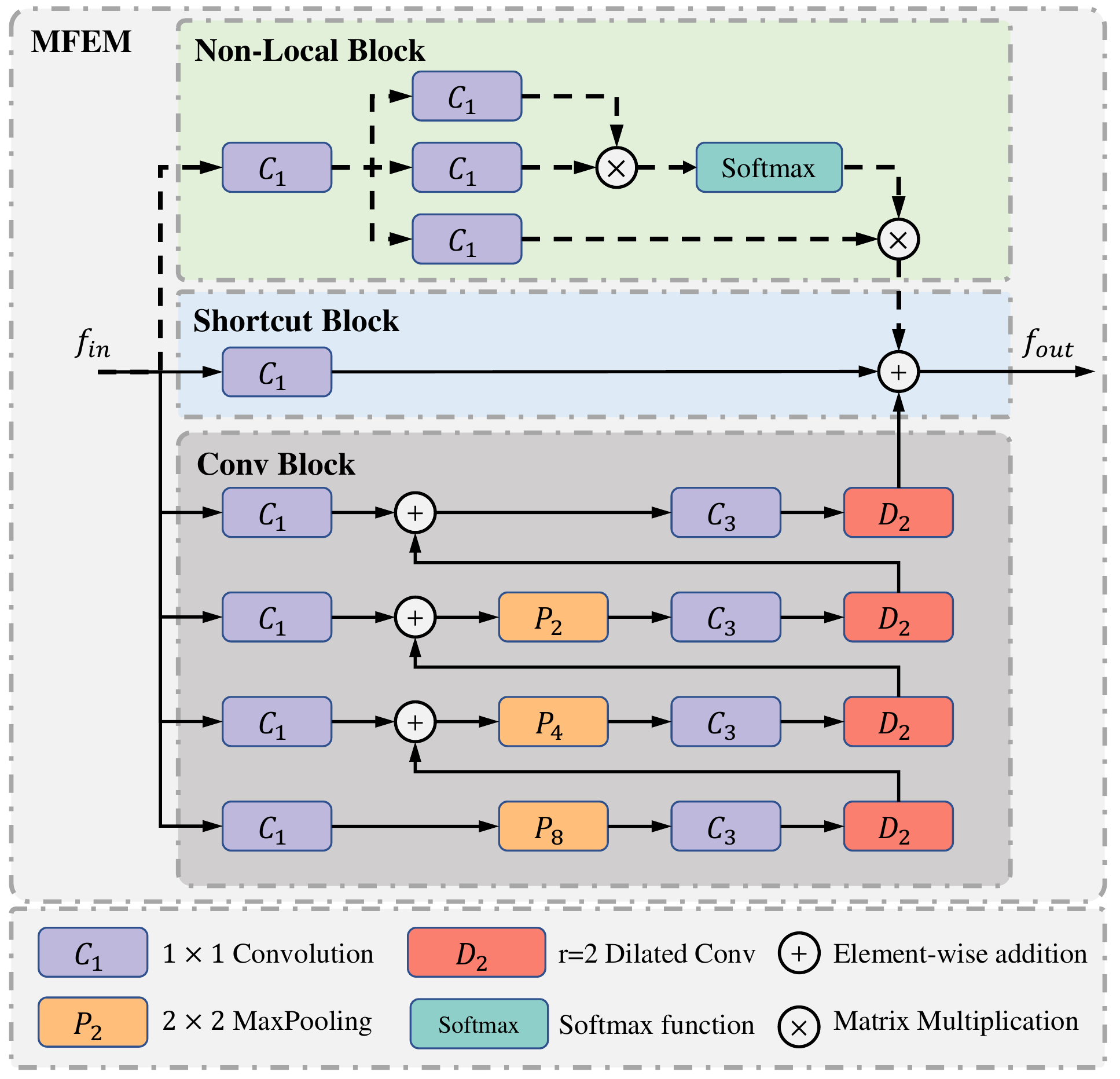}
    \caption{The structure of the MFEM. MFEM contains three blocks, namely Non-local block, shortcut block, and convolutional block. The Non-local block becomes effective only when processing the deepest features (i.e., $f_5$, $F_5$). Upsampling operations are omitted in the figure for conciseness.}
    \label{fig:mfem}
\end{figure}

\subsection{Multi-scale future enhancement module}

The structure of our proposed MFEM is illustrated in \cref{fig:mfem}. As demonstrated in the figure, MFEM consists of three blocks, namely Non-local block, shortcut block, and convolutional block. The three blocks are used in parallel. The output features are then added together via element-wise addition.
 
Specifically, in the Non-local block, we first use a $1\times 1$ convolutional layer to reduce the channel number to 64. Then, three $1\times 1$ convolutional layers are utilized to generate the corresponding key feature $K\in\mathbb{R}^{16\times h\times w}$, query feature $Q\in\mathbb{R}^{16\times h\times w}$, and the value feature $V\in\mathbb{R}^{64\times h\times w}$, where $h$ and $w$ are the height and width, respectively. Afterwards, $K$ and $Q$ are reshaped to ($16\times n$), and $V$ is reshaped to $64 \times n$, where $n=h\times w$. We conduct the matrix multiplication between the transpose of $Q$ and $K$. Afterward, a softmax layer is applied to calculate the spatial attention map $S\in\mathbb{R}^{n\times n}$. Meanwhile, we perform a matrix multiplication between $V$ and $S$. The resulting feature is reshaped to $64\times h \times w$. Thus, the output of the Non-local block can be calculated as follows:
\begin{equation}
    f_{nl} = C_1(f_{in}),
\end{equation}
\begin{equation}
    Q = C_1(f_{nl}), K = C_1(f_{nl}), V = C_1(f_{nl}), 
\end{equation}
\begin{equation}
    S = softmax(Q^T \otimes K),
\end{equation}
\begin{equation}
    f_{nl}^o = V \otimes S,
\end{equation}
where $C_1$ is a $1 \times 1$ convolutional layer, $f_{in}$ is the input feature, $Q^T$ denotes the transpose of $Q$, $\otimes$ represents the matrix multiplication, $f_{nl}^o$ is the output of the Non-local block. Note that to reduce computational overhead, the Non-local block goes into effect only when processing the deepest low-resolution features (i.e., $f_5$ and $F_5$). 

The convolutional block comprises multiple branches. Within each branch, a $1\times 1$ convolution layer is used to reduce the channel count to 64, akin to the approach employed in the Non-local block. As illustrated at the bottom of \cref{fig:mfem}, the initial branch incorporates an $8\times 8$ average pooling layer, succeeded by a $3\times 3$ convolutional layer to expand the receptive fields. Subsequently, a $3\times 3$ dilated convolutional layer, employing a dilation rate of 2, is employed to further enhance the receptive fields without compromising the resolution of the feature map. The resulting feature is upsampled to match the dimensions of the input in the subsequent branch. The two features are subsequently aggregated through element-wise addition.Similarly, the second branch also encompasses a pooling layer and two convolutional layers to refine the features. It is important to note that in the second branch, a smaller average pooling layer (i.e., $4\times 4$) is utilized. As a result, the second branch effectively expands the receptive fields while preserving finer details. By gradually reducing the size of the pooling layers, MFEM is able to capture multi-scale information without sacrificing structural details. Furthermore, by combining the output of the previous branch with the input of the subsequent branch, the receptive fields can be progressively enlarged, enabling the exploitation of global contextual information. Given that the effective receptive field size of a CNN-based model is smaller than the theoretical value, MFEM addresses this limitation with a marginal increase in parameters. The output of the MFEM can be calculated as follows:
\begin{equation}
    f_c^i = C_1(f_{in}),i\in\{1,2,3,4\},
\end{equation}
\begin{equation}
    O_c^i = \begin{cases}
        D_2(C_3(P_8(f_c^1))),&i=1,\\
        D_2(C_3(P_{2^{4-i}}(f_c^i + O_c^{i-1}))),&i=2,3,4,
    \end{cases}
\end{equation}
where $P_i$ is an $i\times i$ average pooling layer, $C_3$ denotes a $3\times 3$ convolutional layer, $D_2$ indicates a $3\times 3$ dilated convolutional layer with a dilation rate of 2, $O_c^i$ is the output of the $i$-th branch, $O_c^i4$ is the output of MFEM. It is noteworthy that we omit the upsampling process for conciseness.

As pointed out in \cite{ResNet,ResidualLearning}, encoding residual features is easier than encoding original features. Thus, we add a shortcut block to facilitate optimization. The result of MFEM can be formulated as follows:
\begin{equation}
    f_s^o = C_1(f_{in}),
\end{equation}
\begin{equation}
    f_c^o = O_c^4,
\end{equation}
\begin{equation}
    f_{out} = f_s^o + f_c^o + f_{nl}^o,
\end{equation}
where $f_{out}$ is the output of MFEM.

\subsection{Boundary enhancement module}

\begin{figure}[htbp]
    \centering
    \includegraphics[width=1\linewidth]{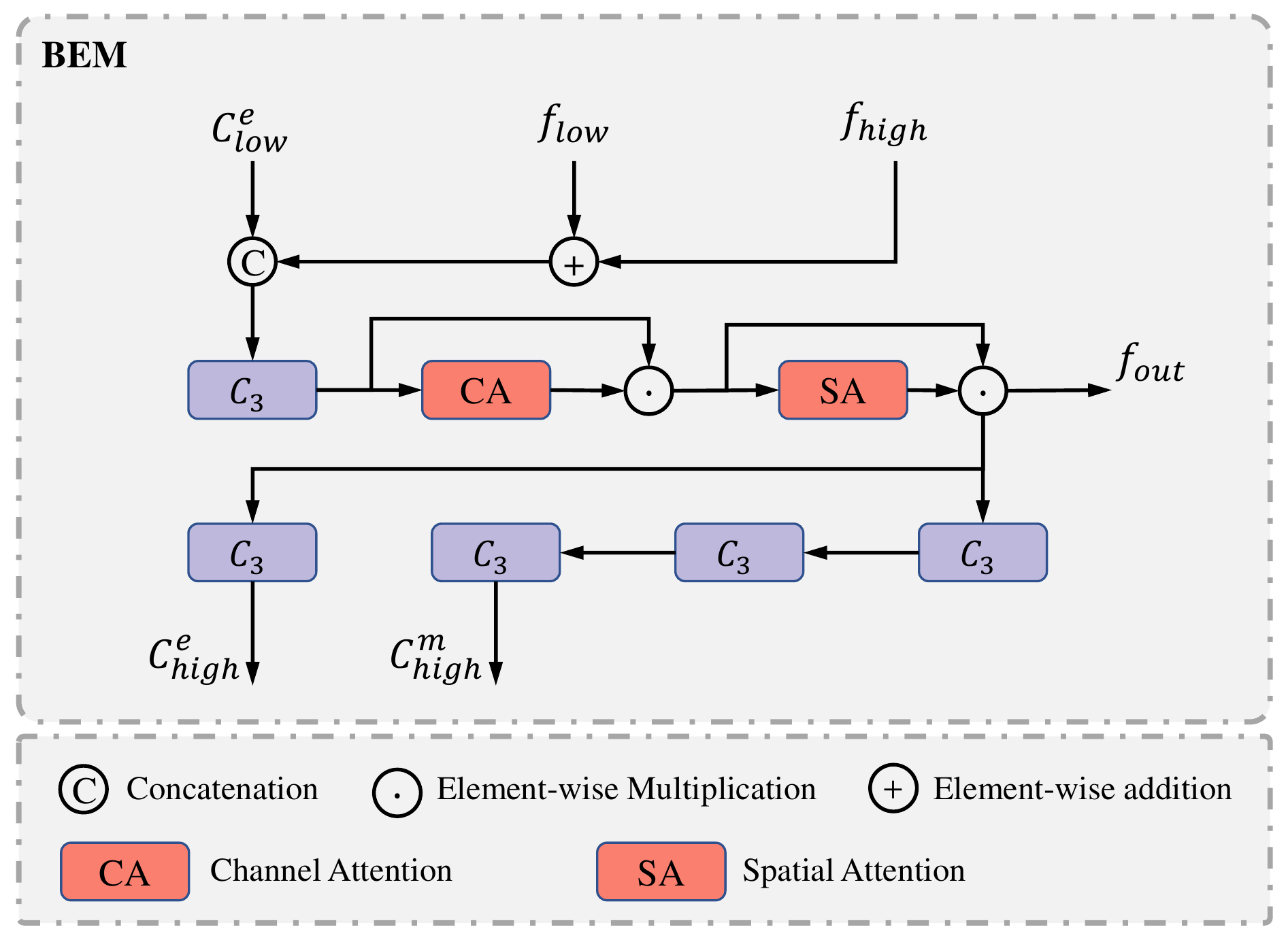}
    \caption{The structure of the BEM.}
    \label{fig:bem}
\end{figure}

The structure of BEM is illustrated in \cref{fig:bem}. $f_{high}$ and $f_{low}$ represent the output features of the former BEM and MFEM respectively. Initially, these two features are integrated using element-wise addition. Subsequently, the resulting feature is combined with the boundary prediction map through channel-wise concatenation. The concatenated feature is then passed through a $3\times 3$ convolutional layer. Considering the boundary prediction map implicitly reveals the location of camouflaged regions, the convolution operation plays a vital role in accurately locating the target. Following this, a channel attention module is employed to emphasize informative channels. As stated in \cite{PFAN}, different channels exhibit responses to different semantics. Hence, incorporating the channel attention operation allows for a more focused analysis of channels related to camouflaged objects. Furthermore, a spatial attention module is utilized to refine the results even further. Based on the refined feature, we generate the mask/boundary prediction maps. The entire process can be formulated as follows:
\begin{equation}
    f =C_3(cat(C_{low}^e, f_{low} + f_{high})),    
\end{equation}
\begin{equation}
    f_{ca} = CA(f)\times f,
\end{equation}
\begin{equation}
    f_{out} = SA(f_{ca}) \times f_{ca},
\end{equation}
\begin{equation}
    C_{high}^e = C_3(f_{out}), C_{high}^m=C_3(C_3(C_3(f_{out}))),
\end{equation}
where $f_{out}$ is the refined feature, $C_{high}^e$ and $C_{high}^m$ are boundary and mask prediction maps, respectively. More concretely, the channel attention module is implemented as:
\begin{equation}
    CA(f) = \sigma(M(P_g(f))),
\end{equation}
where $P_g$ denotes the global average pooling operation, $M$ is a 2-layer perceptron, $\sigma$ represents the sigmoid function. The spatial attention module is implemented as:
\begin{equation}
    SA(f)=\sigma(C_7(P_g^c(f))),
\end{equation}
where $P_g^c$ is a global average pooling operation along the channel dimension, $C_7$ is a $7\times 7$ convolutional layer.

\subsection{Mask-guided fusion module}

\begin{figure}[htbp]
    \centering
    \includegraphics[width=1\linewidth]{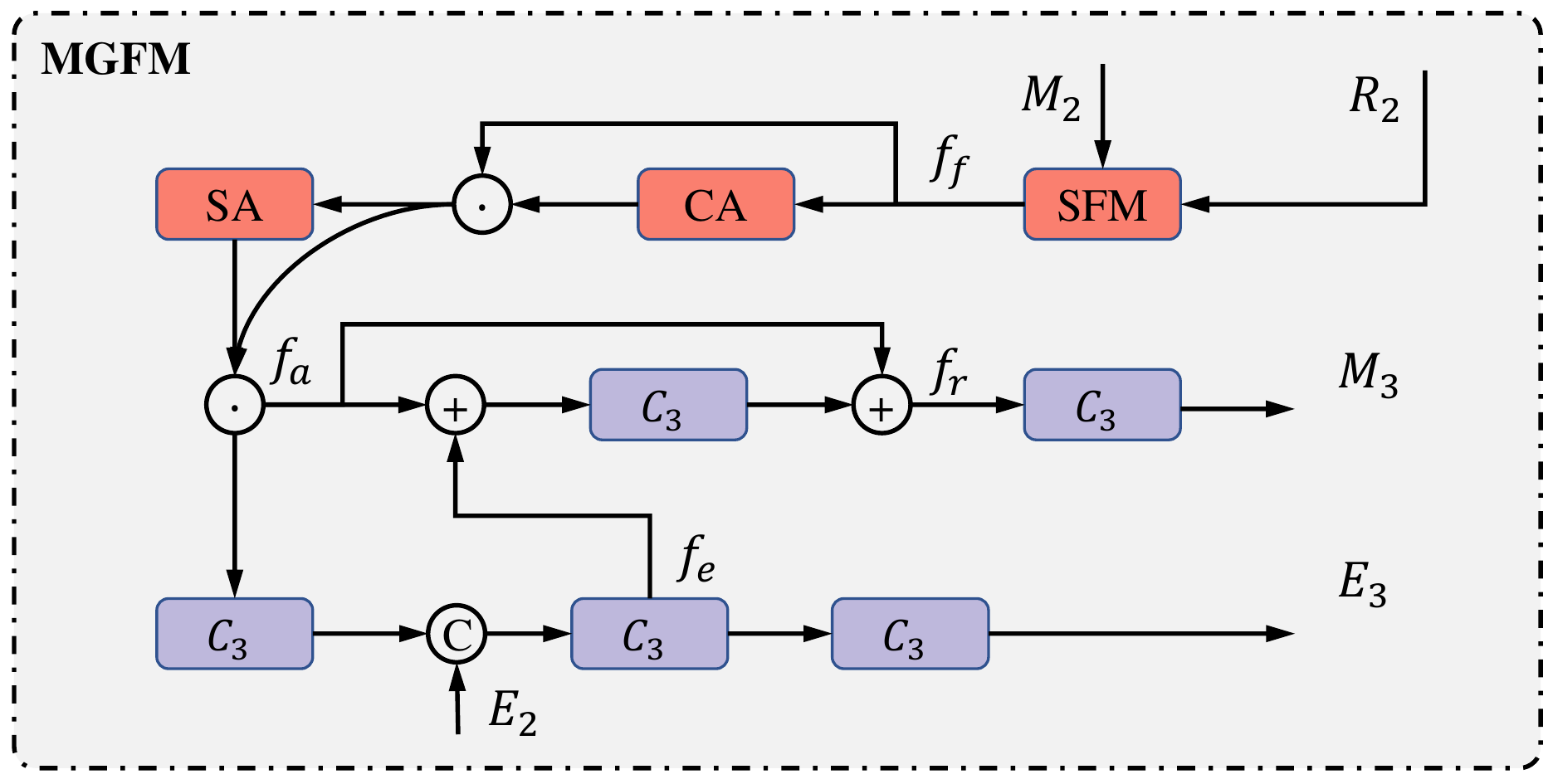}
    \caption{The structure of the MGFM.}
    \label{fig:mgfm}
\end{figure}
\begin{figure}[htbp]
    \centering
    \includegraphics[width=1\linewidth]{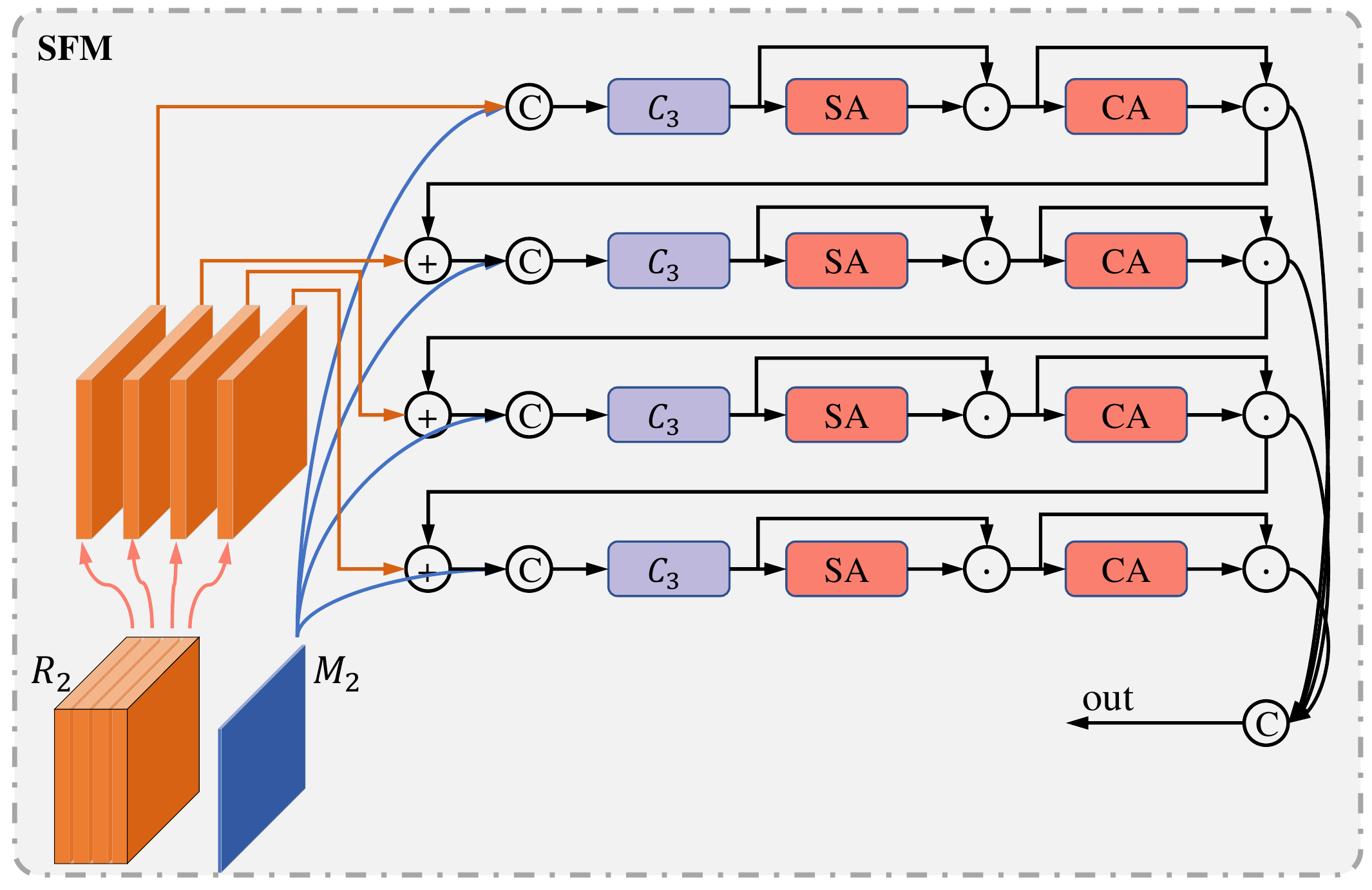}
    \caption{The structure of the SFM. The orange block $f$ and the blue sheet $M$ are respectively the input feature map and mask prediction map.}
    \label{fig:sfm}
\end{figure}

The structure of MGFM is illustrated in \cref{fig:mgfm}. MGFM first uses an SFM to integrate the mask prediction map $M_2$ with the high-resolution feature map $R_2$. The detailed architecture of the proposed SFM is shown in \cref{fig:sfm}. 

Concretely, $f_2$ is inputted into an MFEM to generate $R_2$. The 64-channel feature $R_2$ is divided into four groups, denoted as $\{G^i,i=1,2,3,4\}$. The first sub-feature, which consists of 16 channels, is concatenated with $M_2$ along the channel dimension. The resultant feature is then propagated to a $3\times 3$ convolutional layer. It is noteworthy that the low-level feature $f_2$ contains affluent local details but lacks semantic information. Applying a $3\times 3$ convolutional layer is helpful to eliminate background noise and focus more on foreground regions. Subsequently, a sequential spatial attention operation and a channel attention operation are employed  to further refine the feature. As mentioned in \cite{PFAN}, the spatial attention operation effectively enhances the representation of foreground regions, while the channel attention operation aids in suppressing redundant or noise-degraded channels. Consequently, the resulting feature provides a more accurate description of camouflaged objects.
\begin{figure*}[htbp]
    \centering
    \includegraphics[width=1\linewidth]{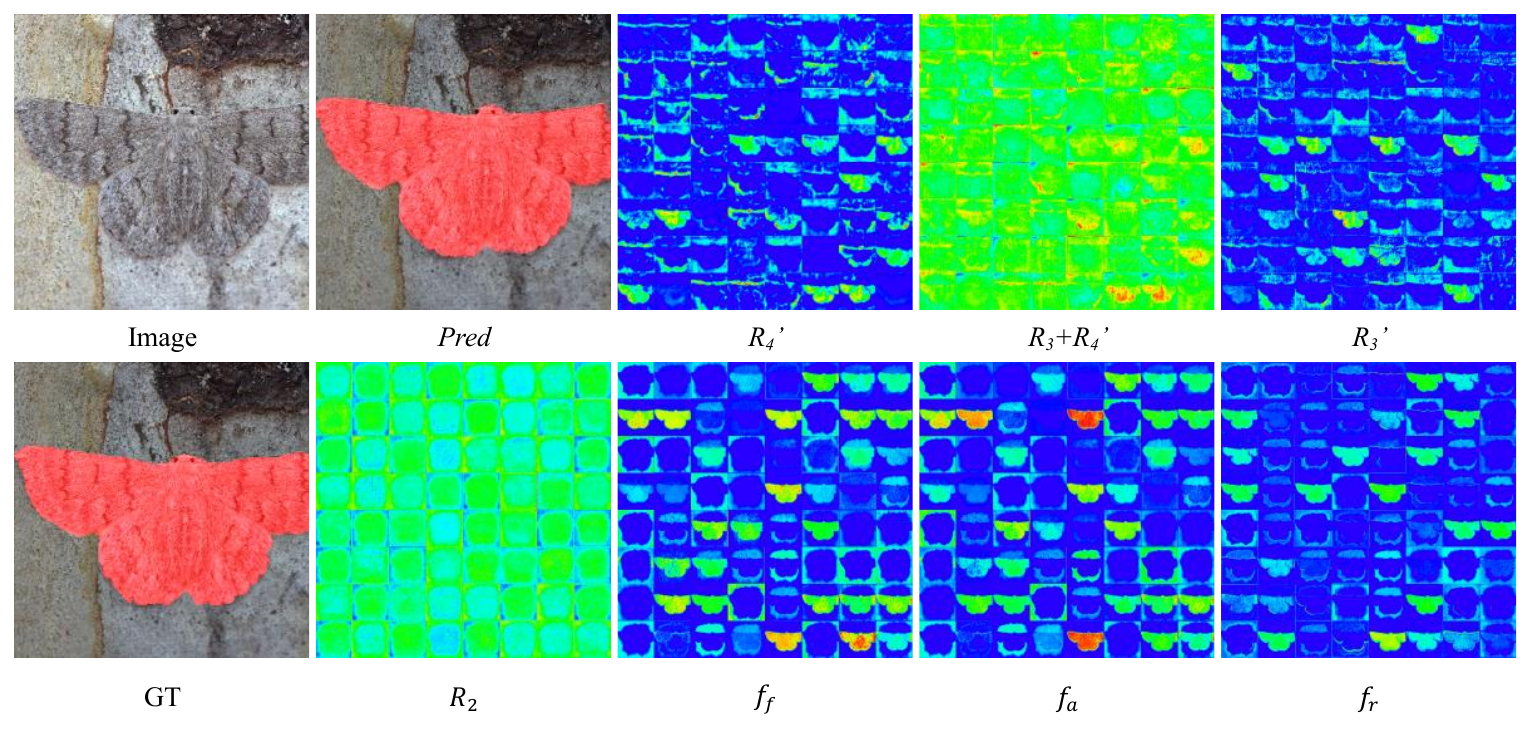}
    \caption{Visualization results. Best viewed in color and zoomed-in.}
    \label{fig:vis-results}
\end{figure*}

The output of the first branch is aggregated with the sub-feature of the subsequent group through element-wise addition. Likewise, we concatenate the aggregated feature with the mask prediction map and employ convolutional and attention operations to further refine it. By iteratively repeating this process, we derive the outputs of the four branches, which are subsequently concatenated to form the output of SFM. This entire process can be formalized as follows:
\begin{equation}
    G^i_{conv} = \begin{cases}
        C_3(cat(G^i,M_2)),&i=1,\\
        C_3(cat(G^i+G^i_o,M_2)),&i=2,3,4,
    \end{cases}
\end{equation}
\begin{equation}
    G^i_{sa}=G^i_{conv}\times SA(G^i_{conv}),
\end{equation}
\begin{equation}
    G^i_o=G^i_{sa}\times CA(G^i_{sa}),
\end{equation}
\begin{equation}
    f_f = cat(G^1_o,G^2_o,G^3_o,G^4_o),
\end{equation}
where $G^i_o$ is the output of the $i$-th branch, $f_f$ is the output of SFM.

Due to the absence of sufficient semantic information, distinguishing between foreground and background regions becomes challenging in $R_2$. While SFM partially addresses this issue, effectively differentiating camouflaged objects from their surroundings remains difficult, leading to potential performance degradation. However, it is worth noting that the boundary prediction map implicitly provides insights into the target's location. Therefore, we propose the introduction of $E_2$ as an enhancement to further improve the performance.

More specifically, we first use attention operations to highlight informative data. 
\begin{equation}
    f_{ca} = f_f \times CA(f_f),
\end{equation}
\begin{equation}
    f_a = f_{ca} \times SA(f_{ca}),
\end{equation}
Then, we calculate the boundary feature by using a single $3\times 3$ convolutional layer. The boundary feature is combined with the boundary prediction map $E_2$ via channel-wise concatenation, the result of which is fed to a $3 \times 3$ convolutional layer for refinement.
\begin{equation}
    f_e = C_3(cat(C_3(f_a), E_2)),
\end{equation}
Thus, we can obtain the boundary-enhanced feature by integrating the boundary feature with the output of the attention operations. Afterwards, we use a shortcut connection to facilitate optimization and employ convolutional layers to compute finer mask/boundary prediction maps. The whole process can be described as:
\begin{equation}
    f_r = f_a + C_3(f_a + f_e),
\end{equation}
\begin{equation}
    M_3 = C_3(f_r),E_3=C_3(f_e),
\end{equation}
where $M_3$ and $E_3$ are respectively mask/boundary prediction maps

To verify the effectiveness of our proposed BEM and MGFM, we present a selection of representative visualization results of feature maps in \cref{fig:vis-results}. The first row of the figure showcases the input features ($R_4' + R_3$) and the output features ($R_3'$) of the final BEM. Additionally, $R_4'$ represents the output feature map of the preceding BEM. Comparing $R_3'$ with $R_4' + R_3$ demonstrates that BEM is effective in distinguishing the foreground regions from the background. Furthermore, from the comparison between $R_3'$ and $R_4'$, we can conclude that BEM is effective in generating features with sharper boundaries. 

Besides, as we can observe from the figure, it is challenging to identify the camouflaged object in $R_2$. A comparison between $R_2$ and $f_f$ illustrates the usefulness of SFM in locating the target and reducing background noise. Furthermore, when compared to $f_f$, it can be observed that certain channels focusing on the background regions (e.g., the third and fifth channels in the first row of $f_f$ and $f_a$) are suppressed in $f_a$, which confirms the beneficial impact of attention operations in highlighting informative channels. Meanwhile, some channels in $f_a$ exhibit difficulty in distinguishing foreground regions from the background (e.g., the seventh channel of the third row and the fourth channel of the last row), which results in fuzzy boundaries that may degrade performance. In contrast, clear contours are observed in $f_r$. Additionally, background channels are further suppressed. The aforementioned comparisons and discussions serve to validate the effectiveness of our proposed modules.

\subsection{loss function}

As done in many previous COD methods \cite{SINetV2,C2FNet,BgNet-KBS,BGNet-IJCAI,FAPNet}, we adopt the hybrid loss function \cite{F3Net} to train the model. The hybrid loss function is defined as:
\begin{equation}
    L = L_{IoU}^w + L_{BCE}^w,
\end{equation}
where $L_{IoU}^w$ is a weighted Intersection-over-Union (IoU) loss, $L_{BCE}^w$ is a weighted Binary Cross Entropy (BCE) loss. As pointed out in \cite{F3Net}, pixels near the boundaries are prone to wrong predictions and should be attached with more importance. The standard BCE and IoU losses ignore the difference between pixels and treat all pixels equally, which results in performance degradation. Differently, $L_{IoU}^w$ and $L_{BCE}^w$ assign larger weights to harder pixels, which has been proven effective in enhancing the model's generalization ability.

We use multiple supervisions for the three side-output maps of the second decoder to facilitate optimization. Thus, the training loss can be formulated as follows:
\begin{equation}
    L_{mask} = L(M_1,G) + \sum_{i=3}^5{L(Rst(C^m_i),G)} + L(M_3,G),
\end{equation}
\begin{equation}
    L_{boundary} = L(E_1,G) + \sum_{i=3}^5{L(Rst(C^e_i),G)} + L(E_3,G),
\end{equation}
\begin{equation}
    L_{total} = L_{mask} + L_{boundary},
\end{equation}
where $G$ denotes the groundtruth, $C^m_i$ and $C^e_i$ are the output mask/boundary prediction maps, $Rst$ denotes the restoring operation to map the prediction map into the original bounding box region, $L_{total}$ is the overall loss function.

\section{Experiments}
\subsection{Datasets and evaluation metrics}

We conduct experiments on three benchmark datasets: CAMO \cite{MirrorNet}, COD10K \cite{SINet}, NC4K \cite{LSR}. The three datasets contain 1,250, 5,066, and 4,121 images, respectively. Following previous COD methods \cite{SINetV2,BgNet-KBS,BGNet-IJCAI,PreyNet,PFNet}, we use 1,000 images from CAMO and 3,040 images from COD10K as our training set. The remaining samples are utilized for evaluation.

To provide a comprehensive evaluation of the performance, we adopt six universally-agreed metrics, including 1) S-measure ($S_m$) \cite{Smeasure}, 2) mean E-measure ($E_\phi$) \cite{emeasure}, 3) weighted F-measure ($F_\beta^w$) \cite{weightedF}, 4) mean absolute error ($M$), 5)Precision-Recall curves, 6) F-measure curves. Note that for $M$, a lower value indicates better performance. For other metrics, higher is better.

\subsection{Implementation details}

We implement BTSNet with the PyTorch toolbox. For fair comparison, we use ResNet-50 \cite{ResNet} to build the bifurcated backbone network. Throughout the training process, all images are resized to $704 \times 704$ and augmented by multiple augmentation strategies (i.e., flipping, rotating, and border clipping). We employ Adam optimizer \cite{Adam} with an initial learning rate of 2e-5 to train the model. We maintain a batch size of 8 during training and adopt the poly strategy with a power of 0.9 for learning rate adjustment. BTSNet is trained for 120 epochs. The training process takes about 15 hours on a single NVIDIA 
GeForce RTX 3090 GPU (24G memory). During testing, the images were resized to $704 \times 704$ as well. For evaluation purposes, we exclusively utilize the output of the third decoder.

\begin{table*}[htbp]
  \centering
  \caption{Comparison of the proposed BTSNet with 17 state-of-the-art algorithms on three COD benchmark datasets. The best results are shown in red. '-' indicates the evaluation results are not available.}
    \begin{tabular}{r|r|c|cccc|cccc|cccc}
    \toprule
    \multirow{2}[2]{*}{Method} & \multicolumn{1}{c|}{\multirow{2}[2]{*}{Pub/Year}} & \multirow{2}[2]{*}{Backbone} & \multicolumn{4}{c|}{CAMO}     & \multicolumn{4}{c|}{COD10K}   & \multicolumn{4}{c}{NC4K} \\
          &       &       & $S_m$    & $F_\beta^w$   & $M$   & $E_\phi$ & $S_m$    & $F_\beta^w$   & $M$   & $E_\phi$ & $S_m$    & $F_\beta^w$   & $M$   & $E_\phi$ \\
    \midrule
    SINet\cite{SINet} & \multicolumn{1}{c|}{CVPR$_{20}$} & ResNet-50 & .751  & .606  & .100  & .771  & .771  & .551  & .051  & .806  & .808  & .723  & .058  & .871  \\
    ERRNet\cite{ERRNet} & \multicolumn{1}{c|}{PR$_{22}$} & ResNet-50 & .747  & .667  & .087  & .849  & .739  & .589  & .048  & .868  & .783  & .704  & .060  & .887  \\
    CubeNet\cite{CubeNet} & \multicolumn{1}{c|}{PR$_{22}$} & ResNet-50 & .788  & .682  & .085  & .838  & .795  & .644  & .041  & .864  & -     & -     & -     & - \\
    TANet\cite{TANet} & \multicolumn{1}{c|}{TCSVT$_{23}$} & ResNet-50 & .778 & .659  & .089  & .813  & .794 & .613 & .043 & .838 & -     & -     & -     & - \\
    TINet\cite{TINet} & \multicolumn{1}{c|}{AAAI$_{21}$} & ResNet-50 & .781 & .678  & .087  & .847  & .793 & .635 & .043 & .848 & -     & -     & -     & - \\
    DTCNet\cite{DTCNet} & \multicolumn{1}{c|}{TMM$_{22}$} & ResNet-50 & .778  & .667  & .084  & .804  & .790  & .616  & .041  & .821  & -     & -     & -     & - \\
    PFNet\cite{PFNet} & \multicolumn{1}{c|}{CVPR$_{21}$} & ResNet-50 & .782  & .695  & .085  & .842  & .800  & .660  & .040  &  .877  & .829  & .745  & .053  & .888  \\
    LSR\cite{LSR}  & \multicolumn{1}{c|}{CVPR$_{21}$} & ResNet-50 & .787  & .696  & .080  & .838  & .804  & .673  & .037  & .880  & .840  & .766  & .048  & .895  \\
    MGL-R\cite{MGL} & \multicolumn{1}{c|}{CVPR$_{21}$} & ResNet-50 & .775  & .673  & .088  & .842  & .814  & .666  & .035  & .890  & .833  & .739  & .053  & .893  \\
    PreyNet\cite{PreyNet} & \multicolumn{1}{c|}{ACMMM$_{22}$} & ResNet-50 & .790  & .708  & .077  & .842  & .813  & .697  & .034  & .881  & .834  & .763  & .050  & .887  \\
    BgNet\cite{BgNet-KBS} & \multicolumn{1}{c|}{KBS$_{22}$} & ResNet-50 & .804  & .719  & .075  & .859  & .804  & .663  & .039  & .881  & .843  & .764  & .048  & .901  \\
    BTSNet &       & ResNet-50 & \textcolor{red}{.824} & \textcolor{red}{.753} & \textcolor{red}{.071} & \textcolor{red}{.875} & \textcolor{red}{.834} & \textcolor{red}{.716} & \textcolor{red}{.033} & \textcolor{red}{.897} & \textcolor{red}{.852} & \textcolor{red}{.781} & \textcolor{red}{.046} & \textcolor{red}{.903} \\
    \midrule
    PraNet\cite{PraNet} & \multicolumn{1}{c|}{MICCAI$_{20}$} & Res2Net-50 & .769  & .663  & .094  & .825  & .789  & .629  & .045  & .861  & .822  & .724  & .059  & .876  \\
    BSANet\cite{BSANet} & \multicolumn{1}{c|}{AAAI$_{22}$} & Res2Net-50 & .794  & .717  & .079  & .851  & .818  & .699  & .034  & .891  & .841  & .771  & .048  & .897  \\
    C$^2$FNet\cite{C2FNet} & \multicolumn{1}{c|}{IJCAI$_{21}$} & Res2Net-50 & .796  & .719  & .080  & .854  & .813  & .686  & .036  & .890  & .838  & .762  & .049  & .897  \\
    BGNet\cite{BGNet-IJCAI} & \multicolumn{1}{c|}{IJCAI$_{22}$} & Res2Net-50 & .813  & .749  & .073  & .870  & .831  & .722  & .033  & .901  & .851  & .788  & .044  & .907  \\
    FAPNet\cite{FAPNet} & \multicolumn{1}{c|}{TIP$_{22}$} &  Res2Net-50 & .769  & .663  & .097  & .802  & .835  & .717  & .034  & .885  & .839  & .753  & .052  & .872  \\
    SINetV2\cite{SINetV2} & \multicolumn{1}{c|}{TPAMI$_{21}$} & Res2Net-50 & .820  & .743  & .070  & .882  & .815  & .680  & .037  & .887  & .847  & .770  & .048  & .903  \\
    BgNet+\cite{BgNet-KBS} & \multicolumn{1}{c|}{KBS$_{22}$} & Res2Net-50 & .832  & .762  & \textcolor{red}{.065} & .884  & .826  & .703  & .034  & .898  & .855  & .784  & .045  & .907  \\
    BTSNet+ &       & Res2Net-50 & \textcolor{red}{.834} & \textcolor{red}{.774} & .066  & \textcolor{red}{.885 } & \textcolor{red}{.854} & \textcolor{red}{.754} & \textcolor{red}{.028} & \textcolor{red}{.913} & \textcolor{red}{.866} & \textcolor{red}{.803} & \textcolor{red}{.040} & \textcolor{red}{.914} \\
    \bottomrule
    \end{tabular}%
  \label{tab:sota}%
\end{table*}%

\begin{table*}[htbp]
  \centering
  \caption{Comparison of the BTSNet with 6 competing methods using $704 \times 704$ images as inputs. The best results are highlighted in red.}
    \begin{tabular}{r|cccc|cccc|cccc}
    \toprule
    \multirow{2}[2]{*}{Method} & \multicolumn{4}{c|}{CAMO}     & \multicolumn{4}{c|}{COD10K}   & \multicolumn{4}{c}{NC4K} \\
          & $S_m$    & $F_\beta^w$   & $M$   & $E_\phi$ & $S_m$    & $F_\beta^w$   & $M$   & $E_\phi$ & $S_m$    & $F_\beta^w$   & $M$   & $E_\phi$ \\
    \midrule
    PFNet\cite{PFNet} & .789  & .709  & .084  & .842  & .827  & .708  & .033  & .893  & .839  & .764  & .050  & .891  \\
    PreyNet\cite{PFNet} & .788  & .715  & .084  & .850  & .831  & \textcolor{red}{.728}  & \textcolor{red}{.031}  & .895  & .840  & .770  & .049  & .891  \\
    BgNet\cite{BgNet-KBS} & .774  & .648  & .106  & .806  & .807  & .641  & .045  & .850  & .827  & .713  & .061  & .863  \\
    BTSNet & \textcolor{red}{.824}  & \textcolor{red}{.753}  & \textcolor{red}{.071}  & \textcolor{red}{.875}  & \textcolor{red}{.834}  & .716  & .033  & \textcolor{red}{.897}  & \textcolor{red}{.852}  & \textcolor{red}{.781}  & \textcolor{red}{.046}  & \textcolor{red}{.903}  \\
    \midrule
    BGNet\cite{BGNet-IJCAI} & .798  & .721  & .081  & .849  & .847  & .745  & .030  & .901  & .849  & .781  & .047  & .894  \\
    BgNet+\cite{BgNet-KBS} & .790  & .668  & .101  & .818  & .826  & .671  & .041  & .867  & .844  & .736  & .057  & .874  \\
    FAPNet\cite{FAPNet} & .769  & .663  & .097  & .802  & .835  & .717  & .034  & .885  & .839  & .753  & .052  & .872  \\
    BTSNet+ & \textcolor{red}{.834} & \textcolor{red}{.774} & \textcolor{red}{.066} & \textcolor{red}{.885} & \textcolor{red}{.854} & \textcolor{red}{.754} & \textcolor{red}{.028} & \textcolor{red}{.913} & \textcolor{red}{.866} & \textcolor{red}{.803} & \textcolor{red}{.040} & \textcolor{red}{.914} \\
    \bottomrule
    \end{tabular}%
  \label{tab:sota-704}%
\end{table*}%

\subsection{Comparisons to the state-of-the-arts}

We compare the proposed BTSNet with 17 state-of-the-art CNN-based models: SINet \cite{SINet}, ERRNet \cite{ERRNet}, CubeNet\cite{CubeNet}, TANet \cite{TANet}, TINet\cite{TINet}, DTCNet\cite{DTCNet}, PFNet\cite{PFNet}, LSR\cite{LSR}, MGL\cite{MGL}, PreyNet \cite{PreyNet}, BgNet \cite{BgNet-KBS}, PraNet \cite{PraNet}, BSANet\cite{BSANet}, C$^2$FNet \cite{C2FNet}, BGNet \cite{BGNet-IJCAI}, FAPNet \cite{FAPNet}, SINetV2\cite{SINetV2}. Since some competing methods (e.g., SINetV2\cite{SINetV2}, PraNet \cite{PraNet}) are built on Res2Net-50 \cite{Res2Net}, we implement BTSNet+  using Res2Net-50 as the backbone for fair comparison.

\begin{figure*}[htbp]
    \centering
    \includegraphics[width=1\linewidth]{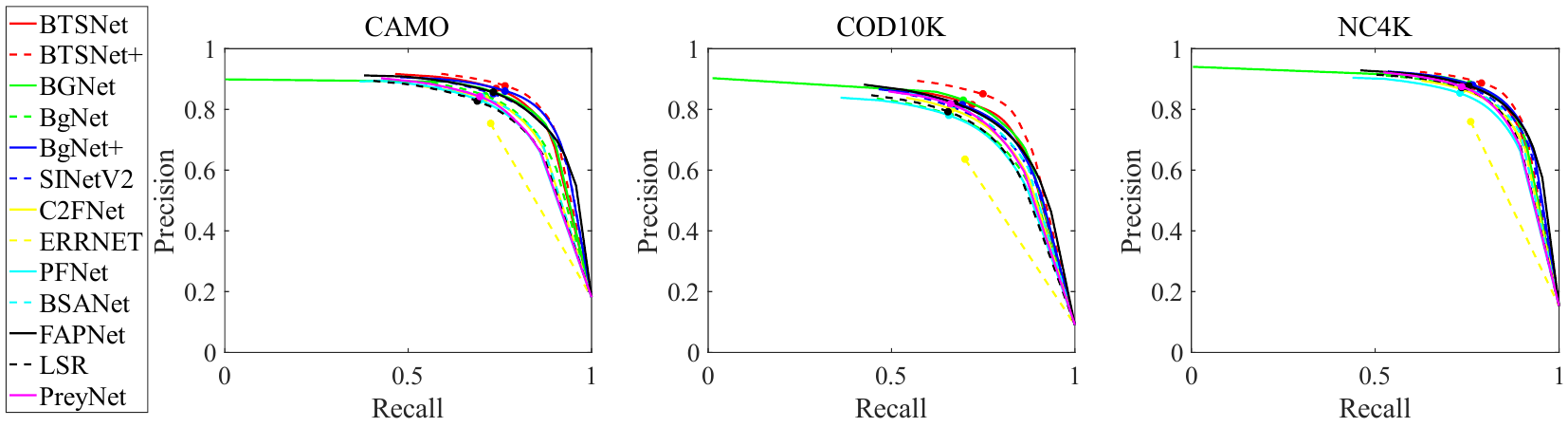}
    \caption{Precison-Recall curves of BTSNet and 11 high-performance COD models.}
    \label{fig:pr}
\end{figure*}

\begin{figure*}[htbp]
    \centering
    \includegraphics[width=1\linewidth]{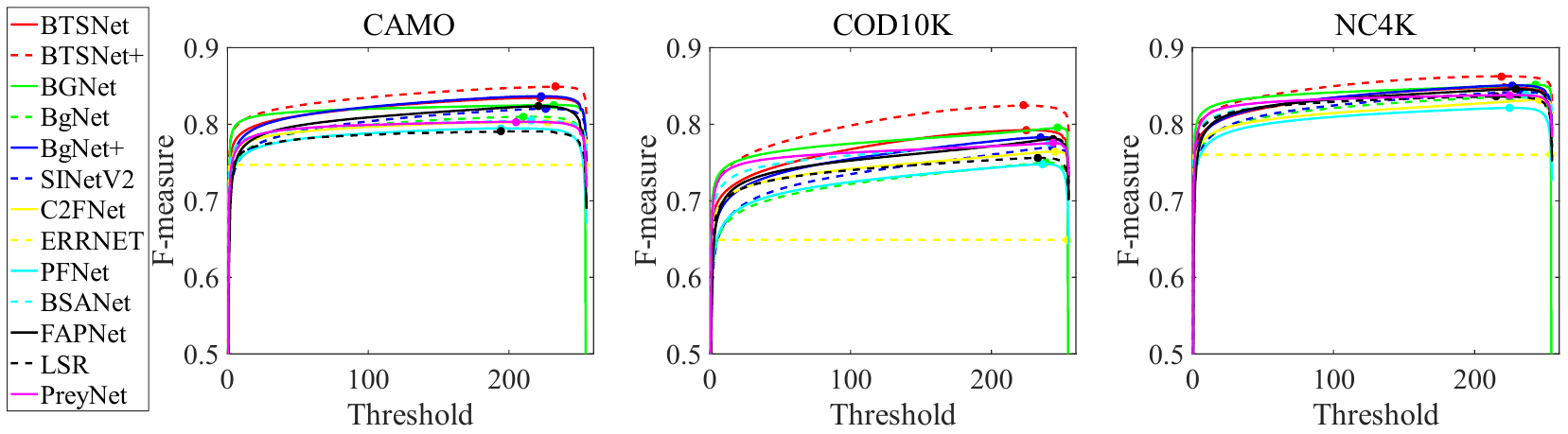}
    \caption{F-measure curves of BTSNet and 11 high-performance COD models.}
    \label{fig:f}
\end{figure*}

\noindent{\textbf{Quantitative Evaluation.}} The quantitative evaluation results of all models are shown in \cref{tab:sota}. It can be clearly seen from the table that BTSNet surpasses other high-performance models across all benchmark datasets in terms of all evaluation metrics. More specifically, performance gains over the three best compared algorithms (PFNet \cite{PFNet}, PreyNet \cite{PreyNet} and BgNet \cite{BgNet-KBS}) built on ResNet-50 are ($S_m: 0.9\%\sim 4.2\%$, $F_\beta^w: 0.9\%\sim5.8\%$, $M: 0.001\sim0.014$, $E_\phi: 0.2\%\sim3.3\%$) on the three challenging datasets. Besides, when using Res2Net-50 as the backbone, BTSNet+ outperforms the three best compared models (BgNet+ \cite{BgNet-KBS}, BGNet \cite{BGNet-IJCAI}, and FAPNet \cite{FAPNet}) by ($S_m: 0.2\%\sim 3.9\%$, $F_\beta^w: 1.2\%\sim7.4\%$, $M: 0\sim0.009$, $E_\phi: 0.1\%\sim2.6\%$). Meanwhile, the Precision-Recall and F-measure curves are shown in \cref{fig:pr} and \cref{fig:f}, respectively. The evaluation results, together with the curves, validate the superiority of the BTSNet.


It is worth noting that the competing methods are trained on images of different sizes. For example, PreyNet\cite{PreyNet}, BGNet\cite{BGNet-IJCAI} and FAPNet\cite{FAPNet} are trained on $448\times 448$, $416\times 416$, and $352 \times 352$ images. Thus, we retrain the six best competing methods (i.e., PFNet\cite{PFNet}, PreyNet\cite{PreyNet}, BgNet\cite{BgNet-KBS}, BGNet\cite{BGNet-IJCAI}, FAPNet\cite{FAPNet}, and BgNet+\cite{BgNet-KBS}) on $704 \times 704$ images for fair comparison. During the training process, according to the Linear Scaling Rule\cite{LSRule}, we reduce the batch size due to the limitation of GPU memory size and adjust the learning rate proportionally. The evaluation results on the three datasets are presented in \cref{tab:sota-704}. As demonstrated in the table, BTSNet still outperforms the six competing methods. Concretely, performance gains over the three models based on ResNet-50 are ($S_m: 0.3\%\sim 5.0\%$, $F_\beta^w: -0.8\%\sim10.5\%$, $M: -0.002\sim0.035$, $E_\phi: 0.2\%\sim6.9\%$). The negative number denotes that a contender (i.e., PreyNet) shows better performance. It is worth noting that BgNet suffers from severe performance degradation, which can be partly attributed to the smaller receptive field size of the model. When using Res2Net-50 as the backbone, BTSNet+ outperforms the three competing algorithms (i.e., BGNet\cite{BGNet-IJCAI}, BgNet+ \cite{BgNet-KBS}, and FAPNet\cite{FAPNet}) by ($S_m: 0.7\%\sim 6.5\%$, $F_\beta^w: 0.9\%\sim11.1\%$, $M: 0.002\sim0.035$, $E_\phi: 1.2\%\sim8.3\%$). 

As we can observe from \cref{tab:sota-704}, the performance of the six competing methods is closer comparable to that of the BTSNet when evaluated on COD10K. This similarity can be partly attributed to the varying proportions of small images in the three datasets.  To quantify this, we define $A_f$ as the area of the foreground region and $A_t$ as the area of the entire image. By selecting images with $A_f/A_t < 1/8, 1/16$, and $1/32$ from the three datasets, we generate three distinct groups of images: Small8, Small16, and Small32, respectively.  The sizes of these image groups are presented in \cref{tab:dataAnalysis}.  Notably, COD10K exhibits a relatively higher proportion of images containing small targets, as indicated by the table.

\begin{table}[htbp]
  \centering
  \caption{The size of the three groups of small images when using different thresholds.}
    \begin{tabular}{r|c|ccc}
    \toprule
     & Threshold & CAMO(250)  & COD10K(2,026) & NC4K(4,121) \\
    \midrule
    Small8  & 1/8  & 101   & 1,523  & 2,137 \\
    Small16 & 1/16 & 40    & 1,043  & 1,080 \\
    Small32 & 1/32 & 11    & 612    & 476 \\
    \bottomrule
    \end{tabular}%
  \label{tab:dataAnalysis}%
\end{table}%

We evaluated BTSNet and the six competing methods on Small8, Small16, and Small32, respectively. The results of the evaluation are presented in \cref{tab:smallresults-704}. In comparison, the performance of the competing methods approaches that of our proposed BTSNet when using a smaller threshold to collect images. However, considering the overall performance, BTSNet outperforms these methods. This indicates that previous methods struggle to generate precise prediction maps when handling high-resolution images due to their limited effective receptive field size. Moreover, a comparison between $X$-704 (e.g., BgNet-704) and $X$ (e.g., BgNet) reveals that models trained on high-resolution images exhibit superior performance when dealing with images containing small camouflaged objects. This can be attributed to the preservation of structural details in high-resolution images, which facilitates the generation of finer prediction maps by the models.

\begin{table*}[htbp]
  \centering
  \caption{Comparison of the BTSNet with competing methods on images with small targets. Small8: images with $A_f/A_t<1/8$. Small16: images with $A_f/A_t<1/16$. Small32: images with $A_f/A_t<1/32$. We use $X$-704 (e.g., BgNet-704) to indicate the model trained on $704\times 704$ images, and utilize $X$ (e.g., BgNet) to represent the original model.}
    \begin{tabular}{r|cccc|cccc|cccc}
    \toprule
    \multirow{2}[2]{*}{Method} & \multicolumn{4}{c|}{Small8(3,761)}   & \multicolumn{4}{c|}{Small16(2,163)}  & \multicolumn{4}{c}{Small32(1,099)} \\
          & $S_m$    & $F_\beta^w$   & $M$   & $E_\phi$ & $S_m$    & $F_\beta^w$   & $M$   & $E_\phi$ & $S_m$    & $F_\beta^w$   & $M$   & $E_\phi$ \\
    \midrule
    BgNet-704\cite{BgNet-KBS} & 0.799  & 0.617  & 0.041  & 0.843  & 0.766  & 0.539  & 0.037  & 0.806  & 0.718  & 0.437  & 0.039  & 0.746  \\
    BgNet\cite{BgNet-KBS} & 0.791  & 0.584  & 0.039  & 0.843  & 0.751  & 0.492  & 0.034  & 0.804  & 0.699  & 0.381  & 0.033  & 0.743  \\
    PFNet-704\cite{PFNet} & 0.825  & 0.694  & 0.027  & 0.893  & 0.797  & 0.626  & 0.023  & \textcolor{red}{0.869}  & \textcolor{red}{0.755}  & 0.530  & \textcolor{red}{0.021}  & \textcolor{red}{0.823}  \\
    PFNet\cite{PFNet} & 0.803  & 0.653  & 0.031  & 0.881  & 0.768  & 0.571  & 0.028  & 0.850  & 0.720  & 0.466  & 0.027  & 0.794  \\
    PreyNet-704\cite{PreyNet} & 0.825  & \textcolor{red}{0.707}  & \textcolor{red}{0.026}  & 0.891  & 0.798  & \textcolor{red}{0.642}  & \textcolor{red}{0.022}  & 0.866  & \textcolor{red}{0.755}  & \textcolor{red}{0.546}  & \textcolor{red}{0.021}  & 0.821  \\
    PreyNet\cite{PreyNet} & 0.815  & 0.689  & 0.026  & 0.886  & 0.785  & 0.616  & 0.022  & 0.861  & 0.736  & 0.507  & 0.021  & 0.811  \\
    TSNet & \textcolor{red}{0.830}  & 0.699  & 0.029  & \textcolor{red}{0.895}  & \textcolor{red}{0.799}  & 0.626  & 0.026  & 0.867  & 0.750  & 0.519  & 0.027  & 0.812  \\
    \midrule
    BgNet+-704\cite{BgNet-KBS} & 0.819  & 0.647  & 0.038  & 0.858  & 0.789  & 0.574  & 0.034  & 0.826  & 0.745  & 0.483  & 0.035  & 0.773  \\
    BgNet+\cite{BgNet-KBS} & 0.812  & 0.621  & 0.035  & 0.858  & 0.778  & 0.539  & 0.030  & 0.826  & 0.731  & 0.435  & 0.027  & 0.771  \\
    BGNet-704\cite{BGNet-IJCAI} & 0.840  & 0.725  & 0.025  & 0.896  & 0.817  & 0.668  & 0.020  & 0.875  & 0.775  & 0.570  & \textcolor{red}{0.018}  & 0.828  \\
    BGNet\cite{BGNet-IJCAI} & 0.831  & 0.713  & 0.026  & 0.903  & 0.802  & 0.646  & 0.022  & 0.880  & 0.761  & 0.552  & 0.021  & 0.834  \\
    FAPNet-704\cite{FAPNet} & 0.826  & 0.688  & 0.028  & 0.871  & 0.799  & 0.623  & 0.024  & 0.844  & 0.758  & 0.534  & 0.023  & 0.795  \\
    FAPNet\cite{FAPNet} & 0.822  & 0.683  & 0.030  & 0.887  & 0.790  & 0.608  & 0.026  & 0.860  & 0.743  & 0.505  & 0.027  & 0.809  \\
    TSNet+ & \textcolor{red}{0.848}  & \textcolor{red}{0.734}  & \textcolor{red}{0.024}  & \textcolor{red}{0.909}  & \textcolor{red}{0.822}  & \textcolor{red}{0.671}  & \textcolor{red}{0.020}  & \textcolor{red}{0.889}  & \textcolor{red}{0.780}  & \textcolor{red}{0.577}  & 0.019  & \textcolor{red}{0.844}  \\
    \bottomrule
    \end{tabular}%
  \label{tab:smallresults-704}%
\end{table*}%

\begin{figure*}[htbp]
    \centering
    \includegraphics[width=1\linewidth]{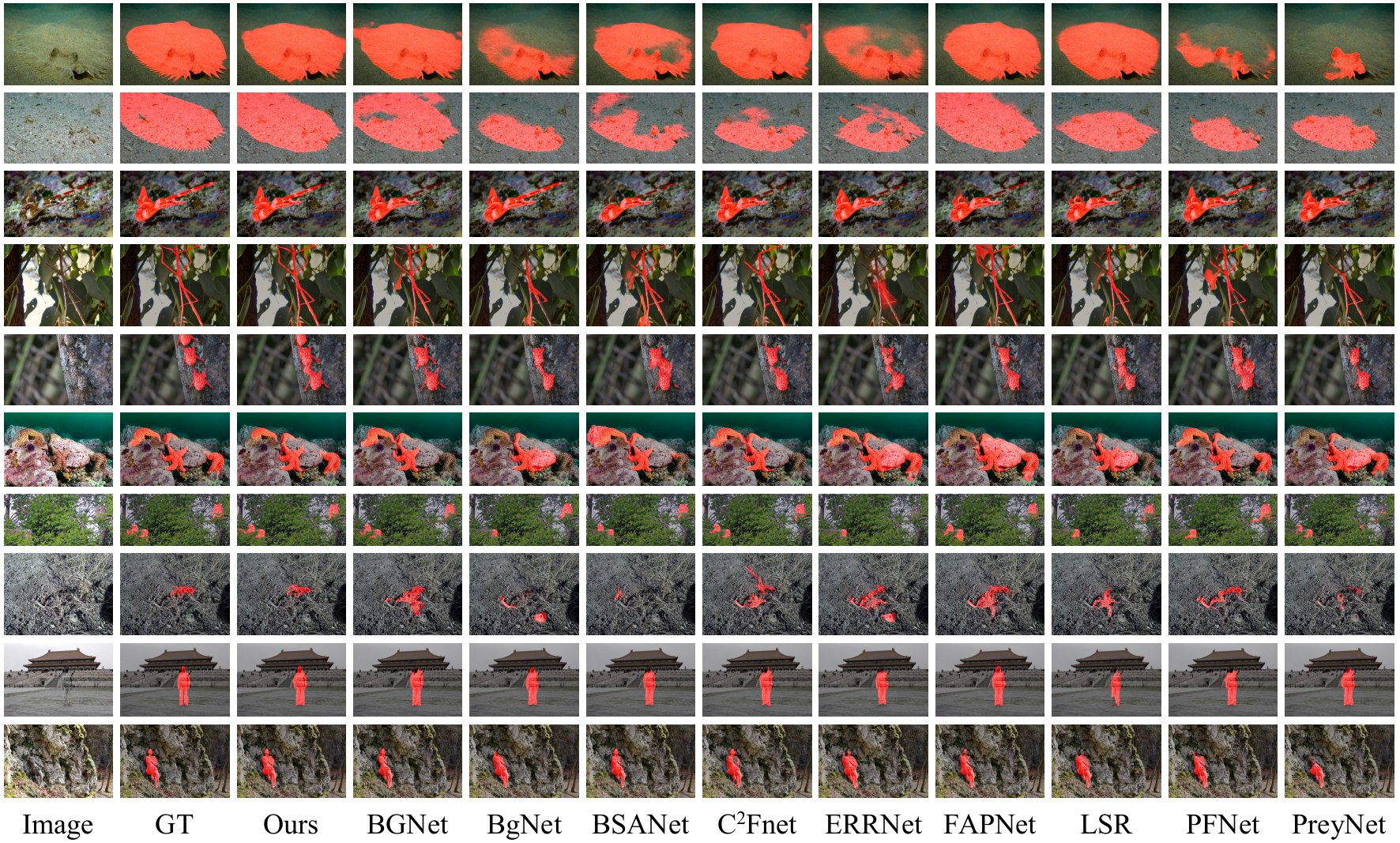}
    \caption{Visual comparisons with 9 high-performance COD models in several representative highly challenging scenarios: large objects (rows 1 and 2), thin and long structures (rows 3 and 4), multiple objects (row 5 and 6), small objects (rows 7 and 8), highly camouflaged regions (rows 9 and 10).}
    \label{fig:sota}
\end{figure*}

\noindent{\textbf{Qualitative Evaluation.}} In \cref{fig:sota}, we visualize several challenging scenes and prediction maps produced by BTSNet and other high-performance models. As we can observe, BTSNet can generate fine-grained results in these scenarios and outperforms other competing methods.

More specifically, the first and second rows illustrate the prediction maps featuring large camouflaged objects. As depicted in the figure, BTSNet effectively captures the entire target, whereas the competing methods exhibit a tendency to overlook certain parts of the objects. The third and fourth rows depict images containing thin and elongated structures. It is evident that the proposed BTSNet accurately segments the camouflaged regions, while the alternative approaches fail to encompass the entirety of the targets. In the fifth and sixth rows, multiple camouflaged objects are present. BTSNet successfully segments all the camouflaged objects, whereas other methods yield results with notably lower accuracy. The seventh and eighth rows showcase images featuring extremely small objects. In comparison with other models, BTSNet not only demonstrates superior precision in target localization but also preserves finer details. The last two rows present challenging scenes characterized by similar foreground and background textures. Although alternative models are capable of locating the primary body of the targets, it is worth noting that they often overlook highly camouflaged regions, such as the head of the camouflaged individuals depicted in the ninth row. In summary, BTSNet proves its competence in generating fine-grained results across a range of highly challenging scenes.

\subsection{Ablation study}

\noindent{\textbf{1 Results of different decoders}}

As shown in \cref{fig:result-decoders}, the results of the third decoder exhibit smoother boundaries. We conducted a quantitative evaluation of the performance of the three decoders, and the results are presented in \cref{tab:decoders}. As illustrated in the table, the first decoder demonstrates the lowest performance, which can be partly attributed to the utilization of low-resolution input features. Furthermore, our primary focus lies in extracting multi-scale information for accurate localization of camouflaged objects, inadvertently overlooking the importance of boundary information. Consequently, harder pixels (e.g., pixels near the boundaries) are prone to erroneous predictions. In general, the third decoder outperforms the second decoder, thereby validating the effectiveness of integrating high-resolution low-level features with the output of the second decoder to enhance performance.

\begin{table*}[htbp]
  \centering
  \caption{Performance of different decoders.}
    \begin{tabular}{r|cccc|cccc|cccc}
    \toprule
    \multirow{2}[2]{*}{Method} & \multicolumn{4}{c|}{CAMO}     & \multicolumn{4}{c|}{COD10K}   & \multicolumn{4}{c}{NC4K} \\
          & $S_m$    & $F_\beta^w$   & $M$   & $E_\phi$ & $S_m$    & $F_\beta^w$   & $M$   & $E_\phi$ & $S_m$    & $F_\beta^w$   & $M$   & $E_\phi$ \\
    \midrule
    First & 0.799  & 0.681  & 0.085  & 0.836  & 0.797  & 0.613  & 0.045  & 0.847  & 0.826  & 0.706  & 0.060  & 0.866  \\
    Second & \textcolor{red}{0.826}  & 0.734  & 0.075  & 0.865  & 0.828  & 0.682  & 0.038  & 0.877  & 0.849  & 0.755  & 0.051  & 0.890  \\
    Third & 0.824 & \textcolor{red}{0.753} & \textcolor{red}{0.071} & \textcolor{red}{0.875} & \textcolor{red}{0.834} & \textcolor{red}{0.716} & \textcolor{red}{0.033} & \textcolor{red}{0.897} & \textcolor{red}{0.852} & \textcolor{red}{0.781} & \textcolor{red}{0.046} & \textcolor{red}{0.903} \\
    \bottomrule
    \end{tabular}%
  \label{tab:decoders}%
\end{table*}%

\noindent{\textbf{2 Effectiveness of the proposed modules}}

\noindent{\textbf{Effectiveness of MFEM.}} To validate the superiority of our proposed MFEM, we conduct several experiments on the three benchmark datasets. Concretely, we train three versions, namely "\textit{w/o} MFEM", "BTSNet-FAM" and "MFEM-Parallel". In "\textit{w/o} MFEM" and "BTSNet-FAM", we replace MFEMs with $3\times 3$ convolution layers and FAMs \cite{PoolNet}, respectively. In "MFEM-Parallel", we change the architecture of the convolutional block. Thus, multiple branches of the block are used in parallel. The quantitative evaluation results are presented in \cref{tab:mfem}. 

More specifically, comparing BTSNet with "\textit{w/o} MFEM" demonstrates that MFEM is effective in largely improving the performance. In contrast to MFEM, FAM\cite{PoolNet} comprises four branches, each employing a pooling layer followed by a convolutional layer to perform convolution operations at varying scales. The output features of the four branches are then aggregated to generate the final output. Although FAM\cite{PoolNet} also adopts the pooling-based strategy, BTSNet surpasses "BTSNet-FAM" by ($S_m: 0.6\%\sim 0.9\%$, $F_\beta^w: 1.3\%\sim2.1\%$, $M: 0.001\sim0.006$, $E_\phi: 0.5\%\sim0.9\%$), highlighting the superiority of MFEM over FAM. The performance gains can be attributed to two main factors: 1) MFEM uses more layers (e.g., dilated convolutional layers and Non-local block) to expand the receptive fields; 2) MFEM employs multiple branches for sequential feature processing, enabling the preservation of detailed structural information. As can be seen from the table, BTSNet outperforms "MFEM-Parallel", further validating the advantage of sequentially processing multi-scale information.

\begin{table*}[htbp]
  \centering
  \caption{Ablation analysis for the proposed MFEM. \textit{w/o} MFEM: replacing MFEM with a single $3\times 3$ convolutional layer. BTSNet-FAM: replacing MFEM with FAM. MFEM-Parallel: employing multiple branches to process features in parallel in the convolutional block.}
    \begin{tabular}{r|cccc|cccc|cccc}
    \toprule
    \multirow{2}[2]{*}{expand\_ratio} & \multicolumn{4}{c|}{CAMO}     & \multicolumn{4}{c|}{COD10K}   & \multicolumn{4}{c}{NC4K} \\
          & $S_m$    & $F_\beta^w$   & $M$   & $E_\phi$ & $S_m$    & $F_\beta^w$   & $M$   & $E_\phi$ & $S_m$    & $F_\beta^w$   & $M$   & $E_\phi$ \\
    \midrule
    \textit{w/o} MFEM & 0.810  & 0.719  & 0.075  & 0.858  & 0.820  & 0.691  & 0.035  & 0.888  & 0.845  & 0.765  & 0.049  & 0.897  \\
    BTSNet-FAM & 0.813  & 0.732  & 0.077  & 0.866  & 0.825  & 0.699  & 0.034  & 0.889  & 0.846  & 0.768  & 0.048  & 0.897  \\
    MFEM-Parallel & 0.821 & 0.748 & 0.073 & 0.870 & 0.828 & 0.705 & 0.034 & 0.891 & 0.848 & 0.773 & \textcolor{red}{0.046} & 0.900 \\
    \midrule
    BTSNet & \textcolor{red}{0.824}  & \textcolor{red}{0.753}  & \textcolor{red}{0.071}  & \textcolor{red}{0.875}  & \textcolor{red}{0.834}  & \textcolor{red}{0.716}  & \textcolor{red}{0.033}  & \textcolor{red}{0.897}  & \textcolor{red}{0.852}  & \textcolor{red}{0.781}  & \textcolor{red}{0.046}  & \textcolor{red}{0.903}  \\
    \bottomrule
    \end{tabular}%
  \label{tab:mfem}%
\end{table*}%

\noindent{\textbf{Effectiveness of BEM.}} We train two version ("\textit{w/o} BEM" and "\textit{w/o} edge") to demonstrate the effectiveness of the proposed BEM. Concretely, "\textit{w/o} BEM" replaces the BEM with a $3\times 3$ convolutional layer, and "\textit{w/o} edge" removes the prediction process of the boundary prediction map. The experimental results are shown in \cref{tab:bem}. As can be clearly seen from the table, "\textit{w/o} edge" shows better performance than "\textit{w/o} BEM", which proves the effectiveness of using attention operations to boost the performance. The comparison between BTSNet and "\textit{w/o} edge" demonstrates that employing the boundary prediction map as auxiliary information is helpful to generate finer results.

\begin{table*}[htbp]
  \centering
  \caption{Ablation analysis for the proposed BEM. \textit{w/o} BEM: replacing BEM with a $3 \times 3$ convolutional layer. \textit{w/o} edge: remove the boundary generation process in BEM.}
    \begin{tabular}{r|cccc|cccc|cccc}
    \toprule
    \multirow{2}[2]{*}{} & \multicolumn{4}{c|}{CAMO}     & \multicolumn{4}{c|}{COD10K}   & \multicolumn{4}{c}{NC4K} \\
          & $S_m$    & $F_\beta^w$   & $M$   & $E_\phi$ & $S_m$    & $F_\beta^w$   & $M$   & $E_\phi$ & $S_m$    & $F_\beta^w$   & $M$   & $E_\phi$ \\
    \midrule
    \textit{w/o} BEM  & 0.807  & 0.724  & 0.077  & 0.864  & 0.823  & 0.692  & 0.036  & 0.887  & 0.846  & 0.765  & 0.049  & 0.897  \\
    \textit{w/o} edge & 0.809  & 0.727  & 0.076  & 0.864  & 0.826  & 0.700  & 0.035  & 0.890  & 0.847  & 0.770  & 0.048  & 0.900  \\
    \midrule
    BTSNet   & \textcolor{red}{0.824}  & \textcolor{red}{0.753}  & \textcolor{red}{0.071}  & \textcolor{red}{0.875}  & \textcolor{red}{0.834}  & \textcolor{red}{0.716}  & \textcolor{red}{0.033}  & \textcolor{red}{0.897}  & \textcolor{red}{0.852}  & \textcolor{red}{0.781}  & \textcolor{red}{0.046} & \textcolor{red}{0.903}  \\
    \bottomrule
    \end{tabular}%
  \label{tab:bem}%
\end{table*}%

\noindent{\textbf{Effectiveness of MGFM.}} We train two versions of BTSNet ("\textit{w/o} SFM" and "\textit{w/o} edge") to investigate the effectiveness of the MGFM. In "\textit{w/o} SFM", the high-resolution feature map and the output of the second decoder are directly concatenated. The concatenated feature is then fed to a $3\times 3$ convolutional layer followed by two attention operations for refinement. Then, the edge prediction map is introduced to generate boundary-enhanced results as done in BTSNet. Similar with BTSNet, "\textit{w/o} edge" uses an SFM to integrate the input feature map with the mask prediction map. The only difference is that "\textit{w/o} edge" does not leverage the boundary prediction map. The experimental results are presented in \cref{tab:mgfm}. As we can observe, BTSNet outperforms the two versions. Performance gains are ($S_m: 0.5\%\sim 1.7\%$, $F_\beta^w: 1.1\%\sim2.9\%$, $M: 0.002\sim0.006$, $E_\phi: 0.3\%\sim1.1\%$). Thus, we can conclude that by employing SFM and introducing boundary information, MGFM is beneficial for boosting COD performance.

\begin{table*}[htbp]
  \centering
  \caption{Ablation analysis for MGFM. \textit{w/o} SFM: replacing SFM with a $3\times 3$ convolutional layer. \textit{w/o} edge: removing the edge generation process in MGFM.}
    \begin{tabular}{r|cccc|cccc|cccc}
    \toprule
    \multirow{2}[2]{*}{} & \multicolumn{4}{c|}{CAMO}     & \multicolumn{4}{c|}{COD10K}   & \multicolumn{4}{c}{NC4K} \\
          & $S_m$    & $F_\beta^w$   & $M$   & $E_\phi$ & $S_m$    & $F_\beta^w$   & $M$   & $E_\phi$ & $S_m$    & $F_\beta^w$   & $M$   & $E_\phi$ \\
    \midrule
    \textit{w/o} SFM & 0.815  & 0.732  & 0.074  & 0.866  & 0.820  & 0.687  & 0.037  & 0.884  & 0.847  & 0.765  & 0.049  & 0.898  \\
    \textit{w/o} edge & 0.806  & 0.721  & 0.075  & 0.863  & 0.822  & 0.697  & 0.036  & 0.889  & 0.844  & 0.767  & 0.048  & 0.899  \\
    \midrule
    BTSNet   & \textcolor{red}{0.824}  & \textcolor{red}{0.753}  & \textcolor{red}{0.071}  & \textcolor{red}{0.875}  & \textcolor{red}{0.834}  & \textcolor{red}{0.716}  & \textcolor{red}{0.033}  & \textcolor{red}{0.897}  & \textcolor{red}{0.852}  & \textcolor{red}{0.781}  & \textcolor{red}{0.046} & \textcolor{red}{0.903}  \\
    \bottomrule
    \end{tabular}%
  \label{tab:mgfm}%
\end{table*}%

\noindent{\textbf{3) Other settings}}

\noindent{\textbf{Impact of different sizes of training images.}} On the one hand, utilizing larger-sized images as input allows for the preservation of finer, more intricate structures. On the other hand, this approach presents limitations in terms of the effective receptive field size. Consequently, the model faces challenges in capturing global contextual information when the input images are excessively large. To investigate the impact of training image sizes and determine the optimal size, extensive experiments are conducted. Specifically, multiple versions of the model are trained by varying the size of the training images while maintaining consistent configurations. It is important to note that, for each model, the image size used during the testing phase corresponds to that employed in the training phase. The results are shown in \cref{tab:inputsize}. From the table, we observe that increasing the input image size allows for the incorporation of more detailed structural information while still exploiting global contextual understanding. Notably, the rate of improvement diminishes at higher resolutions and reaches a plateau when the input size reaches $704\times 704$. Consequently, we adopt $704 \times 704$ as the size for the training samples.

\begin{table*}[htbp]
  \centering
  \caption{Ablation analysis for the spatial resolution of the input image. The performance increases as the input size grows, saturating as the input size reaches $704\times 704$}
    \begin{tabular}{r|cccc|cccc|cccc}
    \toprule
    \multirow{2}[2]{*}{size} & \multicolumn{4}{c|}{CAMO}     & \multicolumn{4}{c|}{COD10K}   & \multicolumn{4}{c}{NC4K} \\
          & $S_m$    & $F_\beta^w$   & $M$   & $E_\phi$ & $S_m$    & $F_\beta^w$   & $M$   & $E_\phi$ & $S_m$    & $F_\beta^w$   & $M$   & $E_\phi$ \\
    \midrule
    480   & 0.812  & 0.735  & 0.074  & 0.867  & 0.820  & 0.689  & 0.035  & 0.884  & 0.846  & 0.869  & 0.049  & 0.897  \\
    512   & 0.817  & 0.744  & 0.072  & 0.879  & 0.822  & 0.695  & 0.035  & 0.887  & 0.847  & 0.771  & 0.048  & 0.898  \\
    544   & 0.822  & 0.749  & 0.071  & 0.872  & 0.815  & 0.696  & 0.035  & 0.885  & 0.849  & 0.772  & 0.048  & 0.899  \\
    576   & 0.816  & 0.744  & 0.072  & 0.875  & 0.818  & 0.702  & 0.036  & 0.892  & 0.851  & 0.775  & 0.047  & 0.900  \\
    608   & 0.822  & 0.752  & 0.071  & 0.880  & 0.829  & 0.704  & 0.035  & 0.890  & 0.850  & 0.778  & 0.048  & 0.899  \\
    640   & 0.811  & 0.731  & 0.075  & 0.861  & 0.831  & 0.710  & 0.035  & 0.892  & 0.849  & 0.779  & 0.048  & 0.898  \\
    672   & 0.820  & 0.750  & 0.072  & \textcolor{red}{0.878}  & 0.832  & 0.714  & 0.034  & \textcolor{red}{0.897}  & 0.851  & 0.780  & 0.047  & 0.901  \\
    704   & \textcolor{red}{0.824}  & \textcolor{red}{0.753}  & \textcolor{red}{0.071}  & 0.875  & \textcolor{red}{0.834}  & \textcolor{red}{0.716}  & \textcolor{red}{0.033}  & \textcolor{red}{0.897}  & 0.852  & \textcolor{red}{0.781}  & \textcolor{red}{0.046}  & 0.903  \\
    736   & 0.823  & 0.752  & 0.072  & 0.872  & 0.833  & 0.713  & \textcolor{red}{0.033}  & 0.896  & \textcolor{red}{0.853}  & \textcolor{red}{0.781}  & \textcolor{red}{0.046}  & \textcolor{red}{0.904}  \\
    \bottomrule
    \end{tabular}%
  \label{tab:inputsize}%
\end{table*}%

\noindent{\textbf{Impact of different sizes of $F_2$.}} Similarly, we also investigate the influence of the $F_2$ size. The experimental results are presented in \cref{tab:secondsize}. As can be clearly seen from the table, the performance increases as the size grows. Besides, the performance saturates at a resolution of $120\times 120$.

\begin{table*}[htbp]
  \centering
  \caption{Ablation analysis for the spatial resolution of features cropped from $f_2$.}
    \begin{tabular}{r|cccc|cccc|cccc}
    \toprule
    \multirow{2}[2]{*}{size} & \multicolumn{4}{c|}{CAMO}     & \multicolumn{4}{c|}{COD10K}   & \multicolumn{4}{c}{NC4K} \\
          & $S_m$    & $F_\beta^w$   & $M$   & $E_\phi$ & $S_m$    & $F_\beta^w$   & $M$   & $E_\phi$ & $S_m$    & $F_\beta^w$   & $M$   & $E_\phi$ \\
    \midrule
    72    & 0.812  & 0.729  & 0.074  & 0.864  & 0.824  & 0.695  & 0.036  & 0.888  & 0.847  & 0.768  & 0.049  & 0.898  \\
    88    & 0.820  & 0.748  & 0.072  & 0.875  & 0.830  & 0.707  & 0.034  & 0.894  & 0.849  & 0.773  & 0.047  & 0.900  \\
    104   & 0.814  & 0.741  & 0.073  & 0.869  & 0.833  & 0.715  & 0.032  & 0.897  & 0.853  & 0.783  & 0.045  & 0.904  \\
    120   & \textcolor{red}{0.824}  & \textcolor{red}{0.753}  & \textcolor{red}{0.071}  & \textcolor{red}{0.875}  & \textcolor{red}{0.834}  & \textcolor{red}{0.716}  & \textcolor{red}{0.033}  & \textcolor{red}{0.897}  & \textcolor{red}{0.852}  & \textcolor{red}{0.781}  & \textcolor{red}{0.046}  & \textcolor{red}{0.903}  \\
    136   & \textcolor{red}{0.824}  & \textcolor{red}{0.753}  & 0.072  & 0.872  & 0.832  & \textcolor{red}{0.716}  & 0.035  & 0.895  & 0.851  & 0.780  & 0.048  & 0.902  \\
    \bottomrule
    \end{tabular}%
  \label{tab:secondsize}%
\end{table*}%

\noindent{\textbf{Impact of different expansion ratios.}} As mentioned in \cref{sec:methods}, following the acquisition of the prediction map generated by the first decoder, the calculation of the initial bounding box is conducted, which is subsequently expanded using an expansion ratio $r$. Based on the adjusted bounding box, features are extracted from $f_2$ and resized to a predetermined resolution. Generally, employing a smaller value of $r$ leads to a reduction in the size of the bounding box, thereby accentuating the foreground regions. Nonetheless, it is possible for the prediction maps produced by the initial decoder to overlook certain segments of the camouflaged targets. Conversely, adopting a larger value of $r$ enables the incorporation of a greater area, facilitating the detection of the complete targets. In order to strike a balance, multiple versions are trained, each utilizing different expansion ratios. The outcomes are illustrated in \cref{tab:expandratio}. As observed, BTSNet achieves the optimal performance with an expansion ratio of $r=1.2$.

\begin{table*}[htbp]
  \centering
  \caption{Ablation analysis for the expansion ratio.}
    \begin{tabular}{r|cccc|cccc|cccc}
    \toprule
    \multirow{2}[2]{*}{ratio} & \multicolumn{4}{c|}{CAMO}     & \multicolumn{4}{c|}{COD10K}   & \multicolumn{4}{c}{NC4K} \\
      & $S_m$    & $F_\beta^w$   & $M$   & $E_\phi$ & $S_m$    & $F_\beta^w$   & $M$   & $E_\phi$ & $S_m$    & $F_\beta^w$   & $M$   & $E_\phi$ \\
    \midrule
    1     & 0.820  & 0.746  & 0.074  & 0.865  & 0.831  & 0.713  & 0.036  & 0.880  & 0.848  & 0.776  & 0.049  & 0.894  \\
    1.2   & \textcolor{red}{0.824}  & \textcolor{red}{0.753}  & \textcolor{red}{0.071}  & \textcolor{red}{0.875}  & \textcolor{red}{0.834}  & 0.716  & 0.033  & \textcolor{red}{0.897}  & \textcolor{red}{0.852}  & \textcolor{red}{0.781}  & \textcolor{red}{0.046}  & \textcolor{red}{0.903}  \\
    1.4   & 0.822  & 0.743  & 0.075  & 0.857  & 0.832  & \textcolor{red}{0.717}  & \textcolor{red}{0.032}  & 0.882  & 0.851  & 0.774  & 0.048  & 0.896  \\
    1.6   & 0.821  & 0.746  & 0.074  & 0.866  & 0.830  & 0.708  & 0.036  & 0.883  & 0.847  & 0.762  & 0.049  & 0.896  \\
    1.8   & 0.820  & 0.745  & 0.073  & 0.867  & 0.828  & 0.710  & 0.037  & 0.880  & 0.844  & 0.761  & 0.048  & 0.897  \\
    \bottomrule
    \end{tabular}%
  \label{tab:expandratio}%
\end{table*}%

\noindent{\textbf{Impact of different backbone architecture.}} The use of a bifurcated backbone network has been previously explored in several methods\cite{CPD,BBSNet,BgNet-KBS}. In these methods, the output prediction map of the first decoder is passed through a holistic attention module, which is utilized to emphasize foreground regions in a low-level feature. Subsequently, the refined feature is fed into the second branch of the backbone network. To compare the two types of bifurcated encoders, we trained a version called "BTSNet-BE2" by employing the encoder from \cite{CPD} for feature extraction. Similar to BTSNet, "BTSNet-BE2" includes a $2\times 2$ pooling layer before the third stage of the encoder, aids in capturing global context information.

The experimental results are presented in \cref{tab:architecture}. As can be seen from the table, BTSNet outperforms "BTSNet-BE2" with notable performance gains ($S_m: 0.5\%\sim 1.3\%$, $F_\beta^w: 0.8\%\sim1.8\%$, $M: 0.001\sim0.003$, $E_\phi: 0.2\%\sim1.1\%$). These results validate the effectiveness of our encoder, primarily due to its ability to effectively eliminate background regions in the second branch. Additionally, the input of the second branch is obtained by cropping a high-resolution feature and resizing it to a fixed resolution of $120\times 120$. Consequently, our encoder is better equipped to handle images with small targets.

\begin{table*}[htbp]
  \centering
  \caption{Ablation analysis for the bifurcated backbone network. BTSNet-BE2: adopting the bifurcated encoder of \cite{CPD,BBSNet} as the backbone network.}
    \begin{tabular}{r|cccc|cccc|cccc}
    \toprule
    \multirow{2}[2]{*}{} & \multicolumn{4}{c|}{CAMO}     & \multicolumn{4}{c|}{COD10K}   & \multicolumn{4}{c}{NC4K} \\
      & $S_m$    & $F_\beta^w$   & $M$   & $E_\phi$ & $S_m$    & $F_\beta^w$   & $M$   & $E_\phi$ & $S_m$    & $F_\beta^w$   & $M$   & $E_\phi$ \\
    \midrule
    BTSNet-BE2 & 0.811  & 0.735  & 0.074  & 0.864  & 0.824  & 0.700  & 0.034  & 0.894  & 0.847  & 0.773  & 0.047  & 0.901  \\
    BTSNet & \textcolor{red}{0.824}  & \textcolor{red}{0.753}  & \textcolor{red}{0.071}  & \textcolor{red}{0.875}  & \textcolor{red}{0.834}  & \textcolor{red}{0.716}  & \textcolor{red}{0.033}  & \textcolor{red}{0.897}  & \textcolor{red}{0.852}  & \textcolor{red}{0.781}  & \textcolor{red}{0.046} & \textcolor{red}{0.903}  \\
    \bottomrule
    \end{tabular}%
  \label{tab:architecture}%
\end{table*}%

\subsection{Failure cases and analyses}

\begin{figure}[htbp]
    \centering
    \includegraphics[width=1\linewidth]{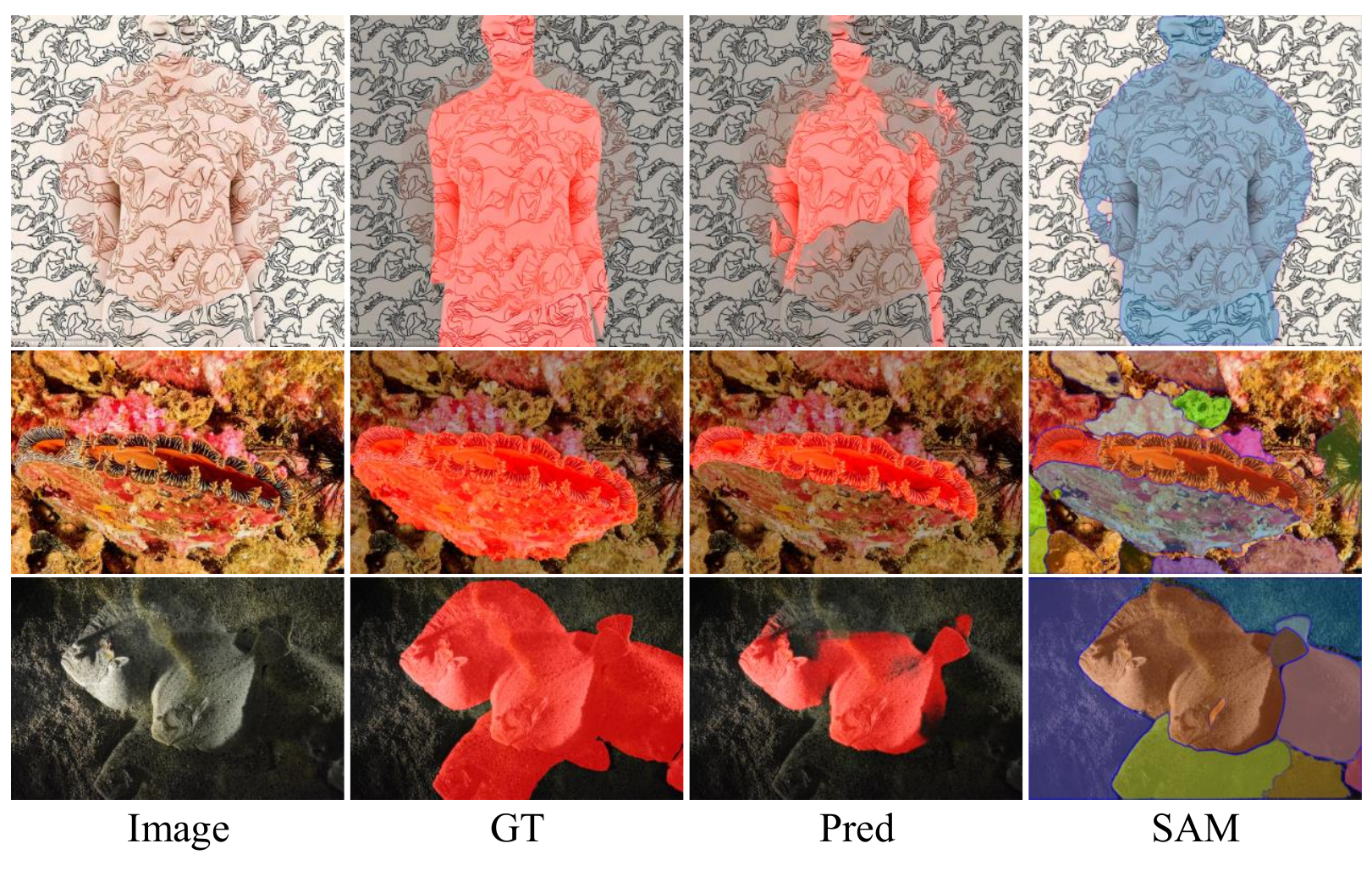}
    \caption{Three representative failure cases of our proposed BTSNet. The results of SAM \cite{SAM} are also presented  for contrast. Best visualized in color and zoomed-in.}
    \label{fig:failurecases}
\end{figure}

Despite the outstanding performance of the proposed BTSNet, it can still exhibit subpar performance in highly challenging scenes. In order to facilitate future research on COD, we present three representative failure cases in \cref{fig:failurecases}. For the purpose of comparison, we also include the results obtained by the recently developed SAM algorithm\cite{SAM}. The first situation is that BTSNet misses some parts of the camouflaged objects. For instance, in the first row, BTSNet fails to accurately detect the camouflaged individuals. It is important to note that although SAM demonstrates better performance, its results are also imperfect. This can be attributed to the fact that the camouflaged target shares the same texture as the background, and the presence of visual noise within the circular region further exacerbates the issue. The second type of failure case occurs when the camouflaged objects are extremely rare. In such scenarios, both BTSNet and SAM struggle to correctly identify the targets or fully segment the objects. In the third category, our model fails to detect targets that are concealed in darkness, which can be partly attributed to the significant illumination variations in the foreground regions. Surprisingly, SAM exhibits much better performance in this particular situation.

We propose several ideas to address the aforementioned instances of failure and anticipate that these suggestions will stimulate insightful considerations for future research in the field of COD. Firstly, we recommend the utilization of a model with larger or even unlimited receptive fields. This proposition is supported by the comparison depicted in \cref{fig:failurecases}, where the performance of SAM surpasses that of BTSNet. Additionally, the advancements in vision transformers \cite{ViT, pvt, SwinTrans} have led to the emergence of transformer-based COD models \cite{UGTR, DTIT}, which exhibit significantly improved performance compared to their CNN-based predecessors. Although transformer-based models typically necessitate more computational resources and pose challenges when applied to high-resolution images, a viable solution could involve the integration of CNN and transformer to construct a hybrid model. Secondly, current approaches allocate equal attention to all pixels during the processing of input images. However, it is worth noting that certain pixels possess a higher level of information, such as the eyes of animals. Consequently, a promising approach would involve first identifying the most informative regions and subsequently examining the surrounding patches in a progressive manner.

\begin{table*}[htbp]
  \centering
  \caption{Computational complexities of BTSNet and 9 state-of-the-art COD models.}
    \begin{tabular}{r|ccccccccccc}
    \toprule
          & SINet & PFNet & PreyNet & BSANet & FAPNet & BGNet & SINetV2 & BgNet & BgNet+ & BTSNet & BTSNet+ \\
    \midrule
    Param(M) & 48.95 & 46.5  & 38.53 & 32.58 & 29.69 & 79.85 & 26.98 & 60.47 & 60.77 & 51.97 & 52.27 \\
    GMACs & 77.69 & 76.04 & 143.48 & 99.82 & 118.75 & 167.38 & 49.1  & 110.74 & 116.01 & 46.19 & 49.79 \\
    Time(s) & 0.0405 & 0.0404 & 0.0744 & 0.0432 & 0.0538 & 0.0455 & 0.0257 & 0.0492 & 0.0556 & 0.0399 & 0.0526 \\
   $\displaystyle{\frac{\rm Time(s)}{\rm Batch}}$ & $\displaystyle{\frac{0.6732}{24}}$ & $\displaystyle{\frac{1.1624}{48}}$ & $\displaystyle{\frac{1.0077}{20}}$ & $\displaystyle{\frac{1.1136}{30}}$ & $\displaystyle{\frac{0.9140}{18}}$ & $\displaystyle{\frac{1.3566}{34}}$ & $\displaystyle{\frac{1.2591}{60}}$ & $\displaystyle{\frac{0.6727}{16}}$ & $\displaystyle{\frac{0.7956}{16}}$ & $\displaystyle{\frac{1.2283}{60}}$ & $\displaystyle{\frac{1.5221}{60}}$ \\[2mm]
    Average(s) & 0.0281 & 0.0242 & 0.0503 & 0.0371 & 0.0508 & 0.0399 & 0.0210 & 0.0420 & 0.0497 & 0.0205 & 0.0254 \\
    \bottomrule
    \end{tabular}%
  \label{tab:complexity}%
\end{table*}%

\subsection{Computational complexity} We compare the computational complexities of BTSNet and 9 state-of-the-art COD methods to further demonstrate the superiority of our proposed model. All experiments are conducted on a single NVIDIA Titan XP GPU, and the inference speed is calculated using $704 \times 704$ images. The inference time is obtained by running the model 100 times and calculating the average value. The experimental results are presented in \cref{tab:complexity}. It is worth noting that we also report the inference time when fully utilizing the GPU. For instance, in the first column, $\frac{0.6732}{32}$ indicates that 32 is the maximum batch size of SINet tested on Titan XP. When using a batch size of 32, the average inference time per iteration is 0.6732s. As observed from the table, the maximum batch size of BTSNet surpasses that of other competing methods, highlighting that BTSNet requires fewer computing resources. Furthermore, the GMACs (Giga Multiply-Accumulates) of BTSNet are also smaller than those of other methods, indicating that BTSNet exhibits higher computational efficiency.

\section{Conclusion}

In this paper, we present a novel bioinspired three-stage model called BTSNet for COD, drawing inspiration from human behavior when observing images that contain camouflaged objects. Our proposed model aims to address the limitations in existing approaches. To achieve this, we first introduce a novel schema and design a bifurcated backbone network, which allows us to effectively utilize detailed structural information while minimizing computational and memory overhead. Additionally, we propose the Multi-scale Feature Enhancement Module (MFEM) to enhance the representation ability of the model. This module improves the model's capability to capture and represent features at different scales. Furthermore, we introduce the Boundary Enhancement Module (BEM) to incorporate boundary information and facilitate the propagation of useful knowledge to the shallower decoder stage. By leveraging the complementary nature of coarse prediction maps and high-resolution low-level features, our model utilizes the Mask-Guided Fusion Module (MGFM) to generate fine-grained prediction maps. Extensive experiments are conducted on three challenging datasets to evaluate the performance of BTSNet. The results demonstrate the superiority of our model compared to 18 state-of-the-art models, as indicated by significant performance improvements across multiple standard evaluation metrics. In conclusion, our proposed BTSNet model, with its innovative three-stage architecture and modules, outperforms existing approaches in the field of Camouflaged Object Detection. The advancements achieved in this research contribute to the further development and application of bioinspired models in computer vision tasks.

\bibliographystyle{IEEEtran}
\bibliography{IEEEabrv,references}

\end{document}